\documentclass[two-column]{article}

\usepackage{arxiv}
\usepackage{amsmath,amsfonts}
\usepackage{algpseudocode}
\usepackage[ruled]{algorithm2e}
\usepackage{array}
\usepackage{textcomp}
\usepackage{stfloats}
\usepackage{verbatim}
\usepackage{graphicx}
\usepackage{cite}
\usepackage{textcomp}
\usepackage{xcolor}
\usepackage[hyphens]{url}
\usepackage{fancyhdr}
\usepackage{hyperref}
\usepackage{subfigure}
\usepackage{makecell}
\usepackage{multirow}
\usepackage{graphicx}

\graphicspath{{FIGURES/}}

\title{BenchMake: Turn Any Scientific Data Set into a Reproducible Benchmark}

\author{ \href{https://orcid.org/0000-0000-0000-0000}{\includegraphics[scale=0.06]{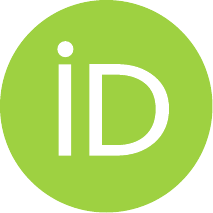}\hspace{1mm}Amanda S.~Barnard}\thanks{https://comp.anu.edu.au/people/amanda-barnard/} \\
	School of Computing\\
	Australian National University\\
	Acton, ACT 2601, Australia \\
	\texttt{amanda.s.barnard@anu.edu.au} \\
}

\date{}

\renewcommand{\shorttitle}{BenchMake}

\hypersetup{
pdftitle={A template for the arxiv style},
pdfkeywords={data set, science, benchmark, archetypal analysis},
}

\begin{document}
\maketitle

\begin{abstract}
Benchmark data sets are a cornerstone of machine learning development and applications, ensuring new methods are robust, reliable and competitive. The relative rarity of benchmark sets in computational science, due to the uniqueness of the problems and the pace of change in the associated domains, makes evaluating new innovations difficult for computational scientists. In this paper a new tool is developed and tested to potentially turn any of the increasing numbers of scientific data sets made openly available into a benchmark accessible to the community.  BenchMake uses non-negative matrix factorisation to deterministically identify and isolate challenging edge cases on the convex hull {(the smallest convex set that contains all existing data instances)} and partitions a required fraction of matched data instances into a testing set that maximises divergence and statistical significance, across tabular, graph, image, signal and textual modalities. BenchMake splits are compared to establish splits and random splits using ten publicly available benchmark sets from different areas of science, with different sizes, shapes, distributions.
\end{abstract}


\section{Introduction}

Benchmark data sets are curated collections that enable consistent, reproducible, and objective evaluation of algorithms and models~\cite{oliver2018realistic, deng2009imagenet}. They are essential for comparing algorithm performance fairly, particularly in machine learning (ML) and artificial intelligence (AI), where the suitability of algorithms can vary widely based on data structure, dimensionality, and distribution~\cite{lecun2015deep, jordan2015machine}. For instance, algorithms that perform exceptionally on structured, tabular data may not generalise well to unstructured image or textual data~\cite{zhang2021understanding}. Established benchmarks such as ImageNet~\cite{deng2009imagenet}, CIFAR data sets~\cite{krizhevsky2009learning}, and OpenML benchmarks for structured data~\cite{vanschoren2014openml} have driven innovation by providing clear metrics for progress, fostering reproducibility and trust within the research community~\cite{Mahmood2025ABC}.

However, in computational sciences, standardised benchmarks remain rare and challenging to establish due to the intrinsic complexity, heterogeneity, and domain specificity of scientific data~\cite{gil2016toward}. Scientific data sets can be represented in a variety of ways (tables, image, text, graphs, signals), often requiring extensive pre-processing, specialised evaluation metrics, and are subject to measurement noise, natural variability, and data imbalance~\cite{wagstaff2012machine}. For example, evaluating models predicting rare diseases necessitates metrics sensitive to false positives and negatives~\cite{bossuyt2003stard}, while climate science prioritises metrics such as the root mean square error (RMSE) for temperature forecasting~\cite{weber2019uncertainty}. Current benchmark examples in science, such as QM9 in quantum chemistry~\cite{ramakrishnan2014quantum}, Wisconsin Breast Cancer data set~\cite{street1993nuclear} and ChestX-ray~\cite{wang2017chestx}, remain exceptions rather than the rule.

A further challenge arises from limitations on data availability due to privacy, ethical, or security considerations~\cite{Mahmood2025ABC,pasquet2019ethical}. Overcoming these barriers and converting scientific data sets into reproducible benchmarks offers considerable value, enhancing transparency, enabling robust performance evaluation, facilitating collaboration, and reducing the costs of repeated data acquisition~\cite{gil2016toward, wagstaff2012machine}. Critically, benchmark data sets must evolve with scientific understanding, distinguishing them from static benchmarks such as handwritten digits (MNIST)~\cite{lecun1998gradient}, housing prices~\cite{harrison1978hedonic}, or celebrity faces~\cite{liu2015deep}.

This paper introduces BenchMake, a novel, deterministic method designed to convert any scientific data set into a benchmark. BenchMake uses non-negative matrix factorization to identify archetypal edge cases~\cite{Javadi2017NonnegativeMF}, partitioning a reproducible sub-set of challenging examples into a test set.  {An ``edge case" refers to an input that lies at the extreme boundary or limits of what a system or program is designed to handle. These cases are unusual, often extreme, and may not be covered by typical testing or usage scenarios~\cite{edge}.} BenchMake handles diverse data types, including tabular data, images, sequential data (such as SMILES strings), graphs and sensor signals, ensuring robust, challenging, and reproducible benchmarking. This enables researchers to publish data sets ready for reproducible benchmarking, facilitating better scientific collaboration and model development.

\section{Methods}

BenchMake is implemented in single point precision {(32 bit floating point precision, FP32)} to improve energy efficiency. Input data is ordered deterministically via stable hashing, and batched to optmise parallelisation. CPU parallelisation is implemented with \texttt{joblib} to improve the computationally intensive distance calculations and batching operations across multiple CPU cores. GPU acceleration is prioritised by first checking for CUDA-capable GPUs via CuPy and closely monitoring the memory usage. BenchMake automatically reverts to CPU-based computations if GPU resources are insufficient or unavailable, as outlined in Alg.\ref{bm}. After processors and data types are selected, BenchMake then moves on to calculate archetypes {residing on the convex hull (the smallest convex set that contains all the points~\cite{Barber1996TheQA})} and identifies the best matching instances in the set to assign as test cases. {As described below, BenchMake is automated so no parameter tuning is required.}

\begin{center}
\begin{minipage}{0.8\textwidth}
\begin{algorithm}[H]
\SetAlgoLined
\KwIn{data set \(X, y\), archetype percentage (test set fraction), data type}
\KwOut{Test set (archetypal instances), Train set (non-archetypal instances)}

Check GPU availability and compatibility; estimate GPU memory requirements\;
\If{GPU resources insufficient}{
    Set computation to CPU mode\;
}
Order data deterministically via stable hashing\;
Determine batch size based on available CPUs\;

\Switch{data type}{
    \Case{tabular}{Load tabular data into numerical format}
    \Case{images}{Load and flatten image data}
    \Case{signal}{Load signals and extract numerical features}
    \Case{sequential}{One-hot encode sequential strings}
    \Case{graph}{Load and numerically encode graph data}
}

Apply Min-Max scaling to data\;

Perform Archetypal Analysis via NMF (Algorithm 2)\;
Calculate distances from data points to archetypes (Algorithm 3)\;
Partition data set into testing (archetypal instances) and training (remaining instances)\;

Return partitioned data sets\;

\caption{Overarching BenchMake Algorithm\label{bm}}
\end{algorithm}
\end{minipage}
\end{center}

\subsection{Non-negative Matrix Factorization}
Non-negative Matrix Factorization (NMF) decomposes a non-negative data set \(X\) into two matrices: a basis matrix \(W\) and a coefficient matrix \(H\), minimizing the reconstruction error~\cite{Lee1999LearningTP}. Compared to Archetypal Analysis (AA)~\cite{cutler1994archetypal,Mrup2010ArchetypalAF,Seth2013ProbabilisticAA}, NMF offers significant computational efficiency and flexibility. AA constrains data points to convex combinations of archetypes, potentially limiting its representation in complex data sets, {whereas NMF does not enforce any constraints.} For data set \(X\) of dimensions \(m \times n\), NMF approximates:
\begin{equation}
X \approx W \times H,
\end{equation}
where \(W\) (dimensions \(m \times k\)) contains weights indicating the contributions of each archetype to the original data points, and \(H\) (dimensions \(k \times n\)) represents the archetypal profiles.  NMF accommodates non-convex distributions through a flexible, interpretable decomposition suited to diverse scientific applications, as shown in Alg.~\ref{nmf}. Unlike Principal Component Analysis (PCA)~\cite{Greenacre2022PrincipalCA}, which allows negative values, NMF maintains non-negativity, enhancing interpretability particularly in domains such as image processing (pixel intensities), text mining (word counts), and bioinformatics~\cite{brunet2004metagenes, pauca2004text}. {NMF is applied globally, and as it is entirely unsupervised, data sets with multiple labels can be handled with ease.}

\begin{center}
\begin{minipage}{0.8\textwidth}
\begin{algorithm}[H]
\SetAlgoLined
\KwIn{Scaled data \(X\), number of archetypes \(k\), max iterations, tolerance}
\KwOut{Matrices \(W, H\) approximating \(X \approx W \times H\)}
Initialize \(W\) (size \(m \times k\)) and \(H\) (size \(k \times n\)) with small random non-negative values\;
\For{iteration = 1 to max iterations}{
    \If{GPU is available}{
        Compute updates for \(H\) using GPU matrix operations:
        \(H \leftarrow H \odot \frac{W^T X}{W^T W H + \epsilon}\)\;
        Compute updates for \(W\) using GPU matrix operations:
        \(W \leftarrow W \odot \frac{X H^T}{W H H^T + \epsilon}\)\;
    }
    \Else{
        Compute updates for \(H\) using CPU matrix operations:
        \(H \leftarrow H \odot \frac{W^T X}{W^T W H + \epsilon}\)\;
        Compute updates for \(W\) using CPU matrix operations:
        \(W \leftarrow W \odot \frac{X H^T}{W H H^T + \epsilon}\)\;
    }
    Enforce non-negativity constraint on \(W, H\)\;
    Compute approximation error \(\|X - W H\|_F^2\)\;
    \If{error change \(\leq\) tolerance}{
        break\;
    }
}
Return matrices \(W, H\)\;
\caption{Archetypal Analysis via NMF\label{nmf}}
\end{algorithm}
\end{minipage}
\end{center}

BenchMake uses \texttt{scipy.optimize.nnls}~\cite{Bro1997AFN} for NMF, iteratively optimizing matrices \(W\) and \(H\) using non-negative least squares (NNLS) to minimize the reconstruction error measured by the Frobenius norm:
\begin{equation}
\min_{W,H} \| X - W \times H \|_F^2.
\end{equation}
This iterative optimization ensures both matrices remain non-negative, identifying interpretable archetypes representing extreme or representative points in the data space~\cite{morch1995archetypal}.  These points do not need to be present in the data set, but could exist in principle.  Real data instances need to be matched to these points using a partitioning scheme.

\subsection{Partitioning Archetypal Edge Cases}
BenchMake selects real-world archetypal instances from the data set to form the benchmark testing set. It computes Euclidean distances between each $i$th data instance and the $j$th archetypes defined by matrix \(H\):
\begin{equation}
D(i, j) = \sqrt{\sum_{k=1}^{n} (X_{i,k} - H_{j,k})^2},
\end{equation}
selecting the closest unique instance for each archetype (in archetypal order).  The top \(k\) instances, determined by a user-defined percentage, form the test set. In the rare event of an exact tie, the instance with the smaller index (in the hashed-and-sorted ordering) is selected as the match for that archetype. This ensures reproducible, challenging benchmarking across diverse data sets without imposing arbitrary distance thresholds, as shown in Alg.~\ref{dist}.

\begin{center}
\begin{minipage}{0.8\textwidth}
\begin{algorithm}[H]
\SetAlgoLined
\KwIn{Scaled data \(X\), archetypes matrix \(H\), batch size}
\KwOut{Distance matrix \(D\) of size \(n_{samples} \times n_{archetypes}\)}

Initialize distance matrix \(D\) to zeros\;
Let \(n_{samples}\) = number of rows in \(X\)\;
Let \(n_{archetypes}\) = number of rows in \(H\)\;

\For{batch start index = 1 to \(n_{samples}\) step batch size}{
    Define batch end index = min(batch start index + batch size - 1, \(n_{samples}\))\;
    Extract data batch \(X_{batch}\) from \(X\)\;
    \If{GPU available}{
        Transfer \(X_{batch}\) and \(H\) to GPU memory\;
        Compute Euclidean distances in parallel on GPU:
        \(D_{batch} = \sqrt{\sum_{features}(X_{batch}[:, newaxis] - H)^2}\)\;
        Transfer computed distances back to CPU memory\;
        Release GPU memory for batch data\;
    }
    \Else{
        Compute Euclidean distances in parallel on CPU:
        \(D_{batch} = \sqrt{\sum_{features}(X_{batch}[:, newaxis] - H)^2}\)\;
    }
    Store \(D_{batch}\) in the appropriate rows of the distance matrix \(D\)\;
}

Return distance matrix \(D\)\;
\caption{Distance Calculations between Data Instances and Archetypes\label{dist}}
\end{algorithm}
\end{minipage}
\end{center}

\subsection{Evaluation Metrics \label{sec:evals}}
Benchmarking is intended to be challenging for an ML or AI method, so an ideal testing set should contain challenging edge cases that are still well aligned to the problem case. Model performance should be reasonable, statistical similarity between the testing and training sets should be low, and distributional divergence {(a measure of dissimilarity)} should be high. To fairly compare BenchMake splitting with alternatives, seven independent evaluations have been used based on 1D histograms of the partitioned training and testing sets (50 bins each):
\begin{enumerate}
    \item \textbf{Kolmogorov–Smirnov (KS) test}: The KS Test is a nonparametric test compares the empirical distribution of a sample against a theoretical distribution (one-sample KS) or compares the distributions of two independent samples (two-sample KS). The KS-test focuses on the maximum distance between the cumulative distribution functions (CDFs), making it sensitive to any shape differences such as shifts, spreads, or tail divergences~\cite{Massey1951TheKT}.
    \item \textbf{T-Test}: The independent t-test calculates a t-statistic based on the difference in group means (for two separate groups), their variances, and sample sizes~\cite{Ross2017}.
    \item \textbf{Mutual Information (MI)}: MI is an information-theoretic measure that quantifies how much the knowledge of one group reduces uncertainty on another. Higher MI means the two groups share more information and are more strongly dependent, meaning lower values indicate a greater difference between the sets~\cite{Kinney2013EquitabilityMI}.
    \item \textbf{Kullback–Leibler (KL) divergence}: The KL divergence is a non-symmetric measure of how one probability distribution \(P\) diverges from a second, reference distribution \(Q\). It represents the amount of additional information required to describe \(P\) when using \(Q\) instead of the true distribution. The KL divergence is always non-negative and is zero only if the two distributions are identical; higher values indicate a greater difference~\cite{Kullback1951OnIA}.
    \item \textbf{Jensen-Shannon (JS) divergence}: The JS divergence is a symmetrical measure of dissimilarity between two probability distributions, constructed by taking the average of each distribution’s KL divergence to their midpoint. It remains finite and is often bounded between 0 and 1 (depending on the chosen log base). A JS divergence of zero indicates that the two distributions are identical; higher values indicate a greater difference~\cite{MENENDEZ1997307}.
    \item \textbf{Wasserstein Distance}: The Wasserstein distance measures the ``cost" of transforming one probability distribution into another, accounting for the geometry of the underlying space, making it particularly useful for continuous distributions. A low Wasserstein distance indicates that the two distributions are more similar; higher values indicate a greater difference~\cite{Villani2009}.
    \item \textbf{Maximum Mean Discrepancy (MMD)}: The MMD is a kernel-based statistical measure to quantify the difference between two probability distributions by comparing their mean embeddings in a reproducing kernel Hilbert space. It returns zero if and only if the distributions are identical; higher values indicate a greater difference~\cite{mmd}. The radial basis function (RBF) kernel is used in this study.
\end{enumerate}

\section{Results and Discussion}

BenchMake is compared against conventional data splitting methods commonly used in practice, including random splits via \texttt{train\_test\_split} and \texttt{TimeSeriesSplit} in Scikit-learn~\cite{pedregosa2011scikit}. Ten different public benchmark data sets, are used including tabular, graph, image, sequential (text) and signal modalities, commonly used to evaluate classification and regression tasks. In each case the train-test split provided in the repository is preserved, and then the sets combined and re-split using {stratified} random sampling (averaged over 20 random seeds) and BenchMake to match the same test sets sizes.  UMAP~\cite{McInnes2018} embeddings {and histograms} are used to visualise the distribution of the testing sets relative to the remaining data in the training sets, to provide qualitative assurance that the testing sets are fairly distributed, and capturing edge cases.

The sections below describe the statistical comparisons of the resultant testing an training sets, using the metrics in Section~\ref{sec:evals}. A sub-set of statistical tests are also used to compare BenchMake to (unstratified) random splits partitioning test set sizes from 0.1 to 0.5. Random forest (RF) model scores are also included; accuracy is used to measure classification performance (higher is better) and the mean squared error (MSE) is used to measure the regression performance (lower is better). The hyperparameters of these models have not been tuned, as the purpose of this exercise is to compare the outputs, not to maximise predictive performance. Nevertheless, it is expectation that BenchMake splits will result in worse performance due to the more challenging edge cases in the testing sub-set.

\subsection{Structured Tables}
The tabular data set included structured numerical and categorical data commonly encountered in typical machine learning benchmarks. Data originated from the UCI Machine Learning Repository OpenML~\cite{OpenML2013}.  For classification, the numeric-only version of the German Credit data set was used, containing  1,000 instances and 20 numerical and features and 2 class labels (``good credit'' and ``bad credit'')~\cite{statlog}, such as status of chequing account, duration of credit, credit history, purpose, credit amount, employment status and personal status. For regression the numeric Boston Housing data set was used, containing 506 instances, 13 features and a continuous target variable indicating house price~\cite{Harrison1978HedonicHP}.  Each data instance corresponds to a town (or census tract) in the Boston area in the 1970s, and the target variable is the median home price in that town. These sets were chosen due to their widespread use as benchmark sets in social science, with established splits. There is no official train-test split provided in these original data sets and so random splitting a test set of 0.2 was used in these examples.

When tabular data is provided (as a NumPy array, Pandas DataFrame, or list), BenchMake first ensures it to a consistent NumPy array so that all numerical operations are performed in float32. Next, it reorders the data rows deterministically by computing a stable hash (using the MD5 algorithm) for each row, to guarantee that the same data produces the same sorted order, regardless of the original row order. BenchMake then applies a Min-Max scaling to the data before partitioning, and returns either four splits (\texttt{X\_train}, \texttt{X\_test}, \texttt{y\_train}, \texttt{y\_test}) in the same data type as the user provided or, if requested, just the lists of indices for the training and testing sets (\texttt{Train\_indices}, \texttt{Test\_indices}).

\begin{table}[ht]
\centering\small
\caption{Comparison of classification and regression results for data splitting methods for the OpenML tabular data sets, representative of social science.  Identical splits sizes are used, and superior results are highlighted in bold.\label{tabular}}
\begin{tabular}{lccc|ccc}
\hline
\multirow{2}{*}{Metric} & \multicolumn{3}{c|}{Classification (German Credit)} & \multicolumn{3}{c}{Regression (Boston Housing)} \\
 & OpenML & Random & BenchMake & OpenML & Random & BenchMake \\
\hline
T-test (p-value)  & 0.5399 & 0.52$\pm$0.0500 & \textbf{0.2884} & 0.5059 & 0.56$\pm$0.0452 & \textbf{0.3691 }\\
KS-test (p-value)   & 0.9444 & 0.93$\pm$0.0143 & \textbf{0.6811} & 0.8457 & 0.88$\pm$0.0309 & \textbf{0.7371} \\
Mutual Information & 0.0017 & 0.0017$\pm$0.0003 & \textbf{0.0066 } & 0.0142  & 0.015$\pm$0.0054 & \textbf{0.0244} \\
KL Divergence  & 0.0769 & 0.05$\pm$0.0148 &\textbf{ 0.5401 } & 0.4011  & 0.4$\pm$0.1149 & \textbf{1.1514} \\
JS Divergence & 0.0375 & 0.04$\pm$0.0031 & \textbf{0.0814}  & 0.0894  & 0.084$\pm$0.0070 & \textbf{0.1221 }\\
Wasserstein Distance  & 0.0268 & 0.029$\pm$0.0029 & \textbf{0.0707 } & 0.0534  & 0.049$\pm$0.0047 & \textbf{0.0806 }\\
Maximum Mean Discrepancy  & 0.0015 & 0.0016$\pm$0.0002 & \textbf{0.0171}  & 0.0056  & 0.0048$\pm$0.0008 & \textbf{0.0138 }\\
\hline
\end{tabular}
\end{table}

\begin{figure}[htbp!] \centering
    \includegraphics[width=1\textwidth]{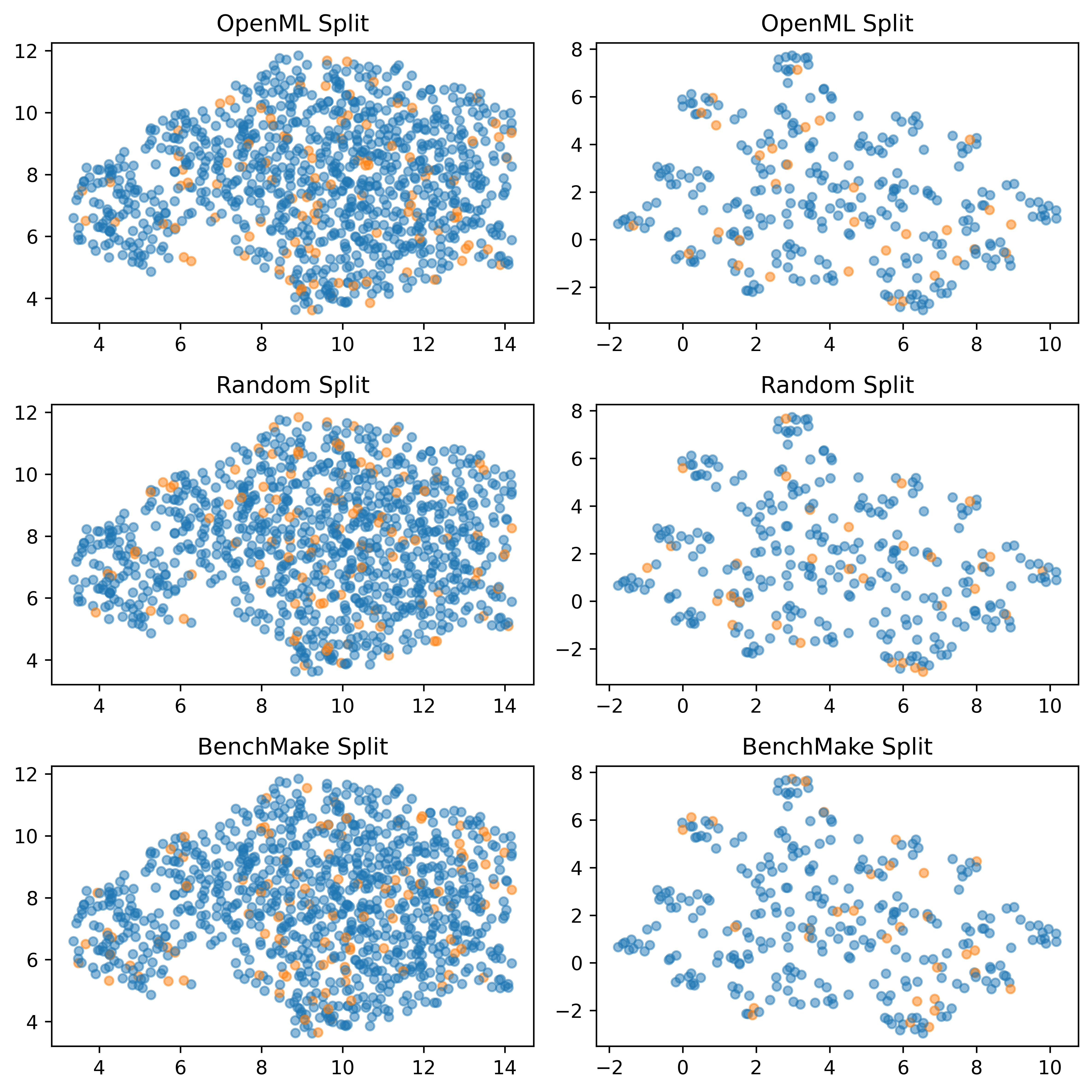}
    \caption{UMAP comparison of the distribution of the training data (blue) and testing data (orange) from the OpenML tabular sets. German Credit set for classification is shown on the left, and Boston Housing for regression is shown on the right.}
    \label{fig:tabular_umap}
\end{figure}

\begin{figure}[htbp!] \centering
    \includegraphics[width=1\textwidth]{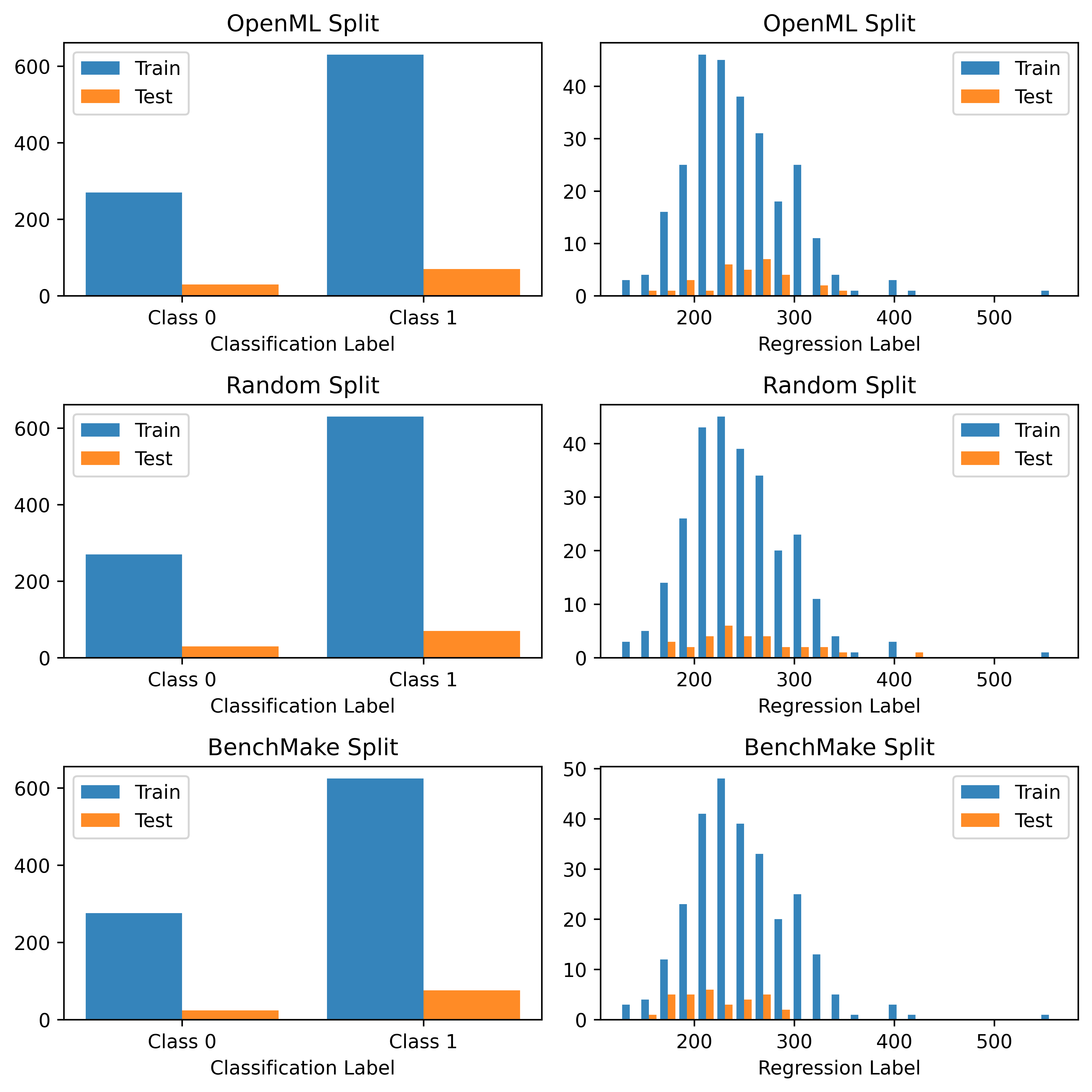}
    \caption{Histogram comparison of the distribution of the training data (blue) and testing data (orange) from the OpenML tabular sets. German Credit set for classification is shown on the left, and Boston Housing for regression is shown on the right.}
    \label{fig:tabular_hist}
\end{figure}

\begin{figure}[htbp!] \centering
    \includegraphics[width=0.8\textwidth]{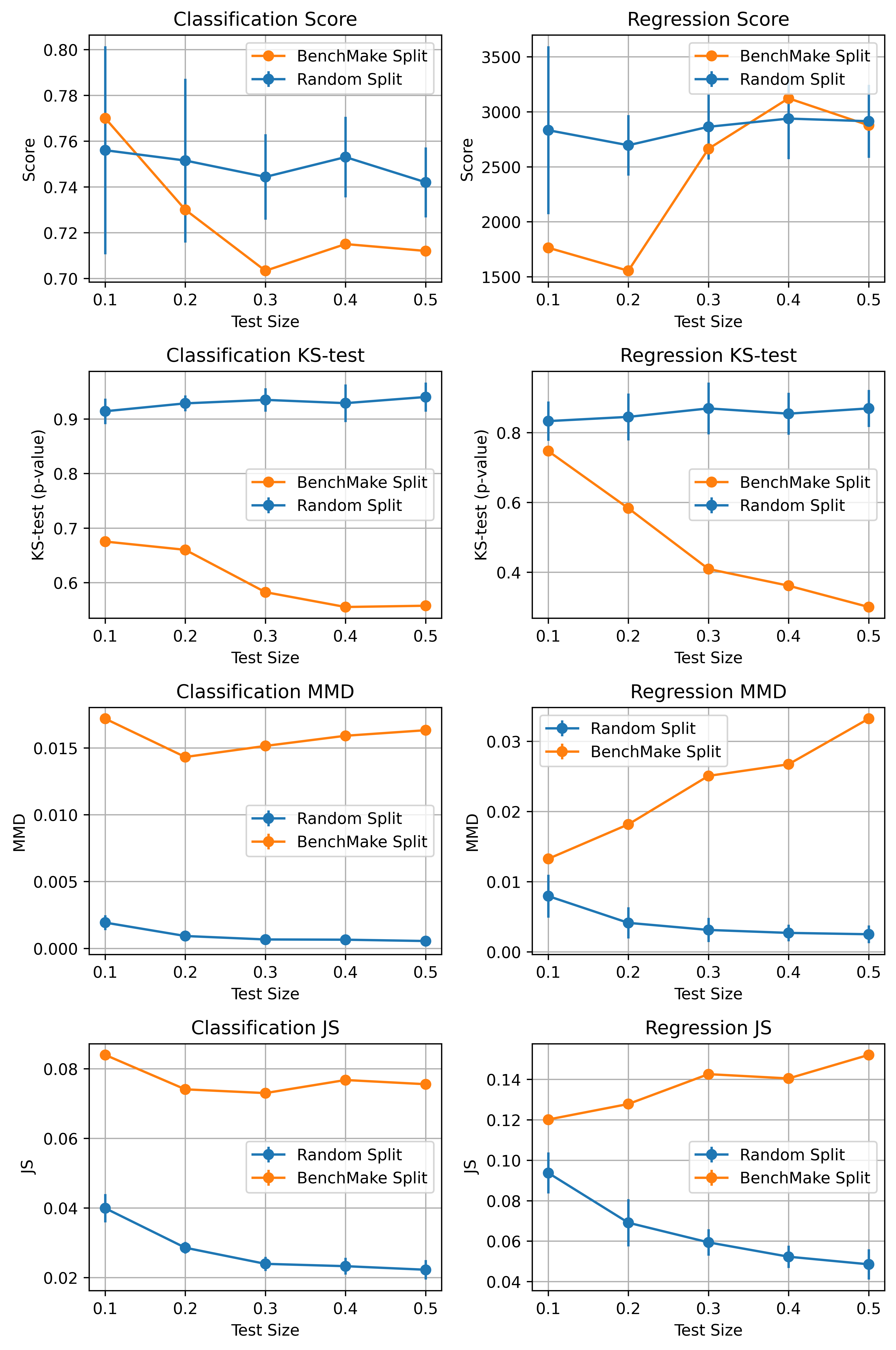}
    \caption{Statistical comparison of the training and testing data for the OpenML tabular sets, split using random \texttt{train\_test\_split} from scikit-learn (blue) and BenchMake (orange). German Credit set for classification is shown on the left, and Boston Housing for regression is shown on the right.}
    \label{fig:tabular_stats}
\end{figure}

The seven statistical test results for the established splits from the OpenML repository, and with the random and BenchMake splits (all with the same test:train ratio), are shown in Table~\ref{tabular}.  In both classification and regression tasks BenchMake generates superior splits, with lower T-test and KS-test p-values, and higher MI, KL and JS divergence, Wasserstein distance and MMD.  In many cases the OpenML splits and the random splits produce similar results; a testament to their random origins. The distributions of the testing and training sets in each case are shown in Fig.~\ref{fig:tabular_umap} and {Fig.~\ref{fig:tabular_hist},} and selected metrics for different BenchMake split sizes are compared to random in Fig.~\ref{fig:tabular_stats}. In Fig.~\ref{fig:tabular_stats} the BenchMake splits become more challenging, divergent and significant as the size of the testing set increases, due to the partitioning or more archetypal edge cases leaving fewer in the training set. The model performance also depends on the splits, both the size and the source {(top row of Fig.~\ref{fig:tabular_stats})}. When over $>$30\% of the data is reserved for testing, the BenchMake partitions are more challenging than random splits for these examples, providing a better test of real-world model performance.

\subsection{Graphs}
The Open Graph Benchmark (OGB) is a collection data sets for machine learning on graphs, facilitating standardized evaluation and scalable research in a number of domains~\cite{ogb}. Among its molecular prediction data sets, MOLHIV is designed for binary classification tasks focused on predicting HIV inhibition~\cite{Wu2017MoleculeNetAB}. In this set each molecular graph is accompanied by a label indicating whether or not it is active against HIV, enabling research into drug discovery and medicinal chemistry. In contrast, the MOLLIPO is a regression data set is designed to predict of lipophilicity, an essential physicochemical property closely tied to the absorption, distribution, metabolism, and excretion (ADME) profile of a drug. MOLLIPO was curated from the ChEMBL database~\cite{Mendez2018ChEMBLTD} and consists of experimental octanol/water distribution coefficients (log D at pH 7.4). MOLHIV contains 41,127 molecules and MOL-LIPO contains 4,200 molecules, and in both data sets atoms are represented by nodes, with a 9-dimensional feature vector (Atomic number, Chirality, Total degree, Formal charge, Total number of H, Number of radical electrons, Hybridization, Is aromatic, Implicit valence) and bonds are represented by edges with a 3-dimensional feature vector (Bond type, Bond stereochemistry, Is conjugated). Instead of randomly splitting the molecules, OGB uses a standard scaffold split based on their molecular scaffold, which is typically defined as the core ring system and the linker atoms connecting them, after removing the side chains.

When provided with graph data, BenchMake assumes a node-feature matrix where each row represents a node and each column represents a feature (this can be in a Pandas DataFrame, NumPy array, or list format). If necessary, the multi-dimensional input is first converted into a two-dimensional float32 array, by flattening any extra dimensions, and stable hashing is applied to the rows to reorder the data. Following Min-Max scaling, BenchMake partitions the data based on the nodes, and the final output will be either the training and testing splits in the same format as the input data or the lists of indices corresponding to these splits.

\begin{table}[ht]
\centering\small
\caption{Comparison of classification and regression results for data splitting methods for the Open Graph Benchmark (OGB) graph data sets, representative of biochemistry. Identical splits sizes are used, and superior results are highlighted in bold.\label{graphs}}
\begin{tabular}{lccc|ccc}
\hline
\multirow{2}{*}{Metric} & \multicolumn{3}{c|}{Classification (MOLHIV)} & \multicolumn{3}{c}{Regression (MOLLIPO)} \\
 & OGB & Random & BenchMake & OGB & Random & BenchMake \\
\hline
T-test (p-value) & 0.6247 & 0.54$\pm$0.0929 & \textbf{0.0000} & 0.2994 & 0.5$\pm$0.1213 & \textbf{0.0000} \\
KS-test (p-value) & 0.5501 & 0.6$\pm$0.1254 & \textbf{0.0100 }& 0.5287 & 0.67$\pm$0.0960 & \textbf{0.0003} \\
Mutual Information & 0.0001 & 0.0003$\pm$0.0001 & \textbf{0.0597} & 0.0016 & 0.0011$\pm$0.0005 & \textbf{0.0603}\\
KL Divergence & 0.0096 & 0.01$\pm$0.0014 & \textbf{2.9858}& 0.1052 & 0.14$\pm$0.0249 & \textbf{3.5735} \\
JS Divergence  & 0.0007 & 0.0007$\pm$0.0001 &\textbf{0.1142} & 0.0064 & 0.0070$\pm$0.0006 & \textbf{0.1231} \\
Wasserstein Distance  & 0.0013 & 0.0016$\pm$0.0003 & \textbf{0.0980} & 0.0058 & 0.0050$\pm$0.0008 &\textbf{ 0.0806} \\
Maximum Mean Discrepancy & 0.0005 & 0.0009$\pm$0.0004 & \textbf{0.0927} & 0.0051 & 0.0031$\pm$0.0009 & \textbf{0.0760 }\\
\hline
\end{tabular}
\end{table}

\begin{figure}[htbp!] \centering
    \includegraphics[width=1\textwidth]{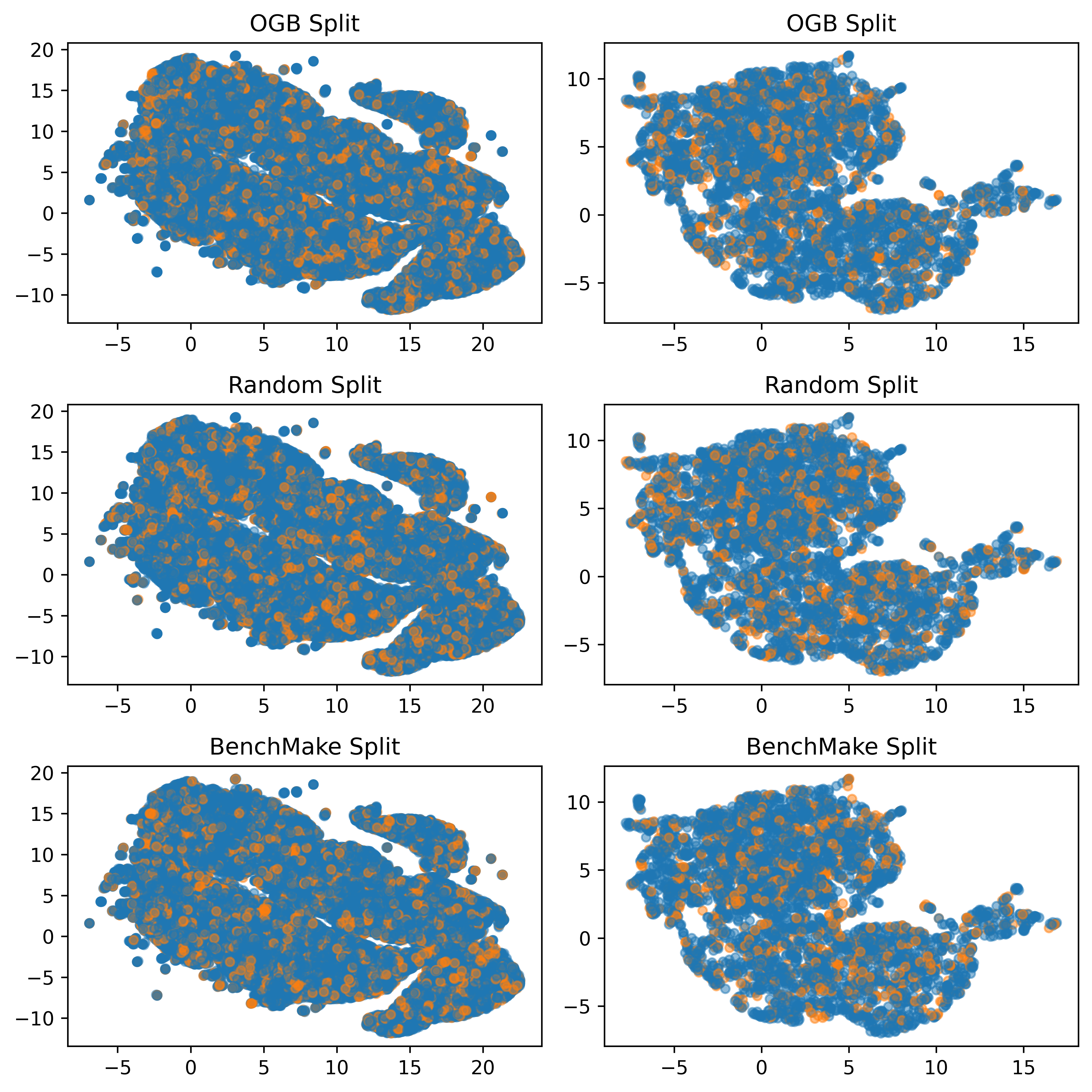}
    \caption{UMAP comparison of the distribution of the training data (blue) and testing data (orange) from the OGB graph sets. MOLHIV set for classification is shown on the left, and MOLLIPO for regression is shown on the right.}
    \label{fig:graph_umap}
\end{figure}

\begin{figure}[htbp!] \centering
    \includegraphics[width=1\textwidth]{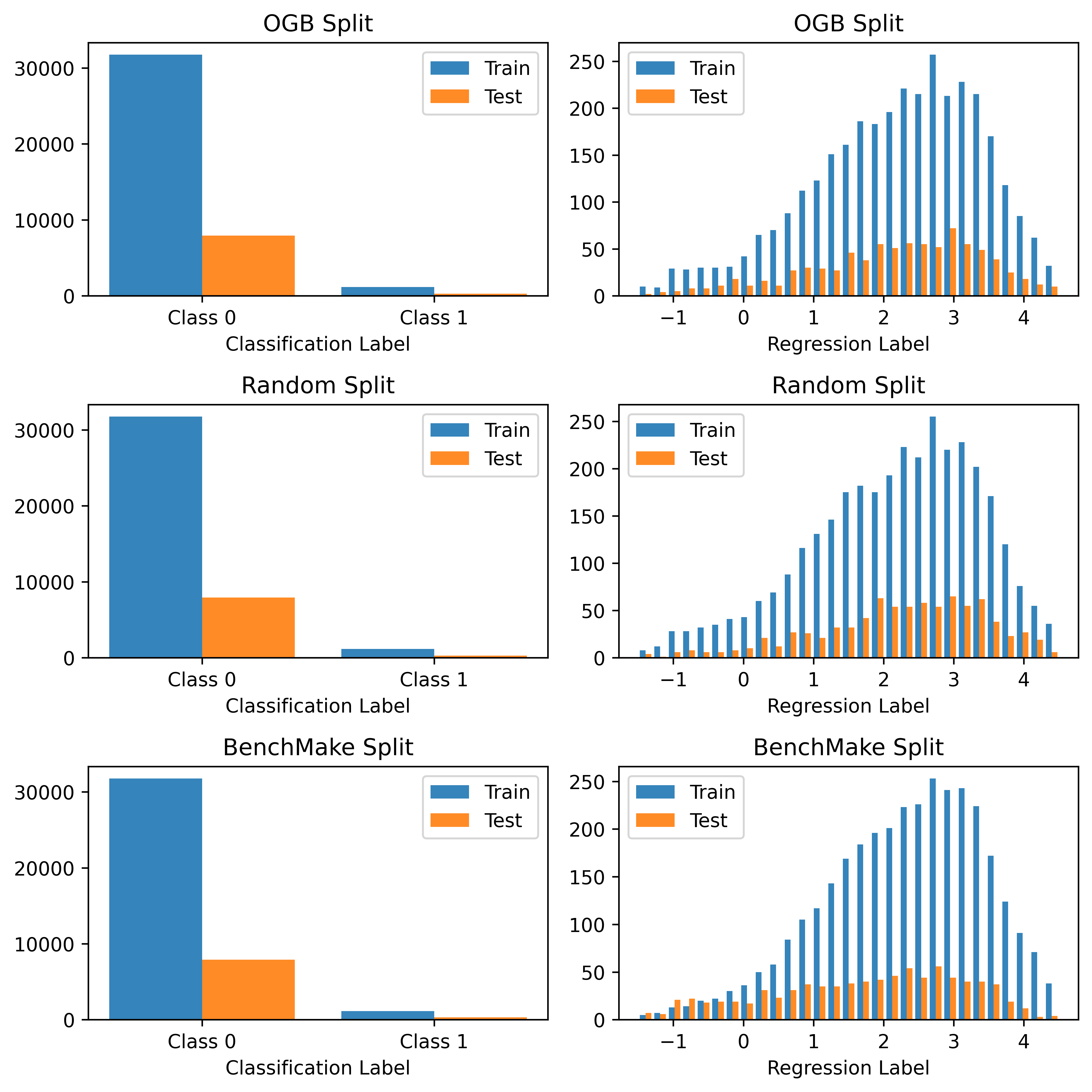}
    \caption{Histogram comparison of the distribution of the training data (blue) and testing data (orange) from the OGB graph sets. MOLHIV set for classification is shown on the left, and MOLLIPO for regression is shown on the right.}
    \label{fig:graph_hist}
\end{figure}

\begin{figure}[htbp!] \centering
    \includegraphics[width=0.8\textwidth]{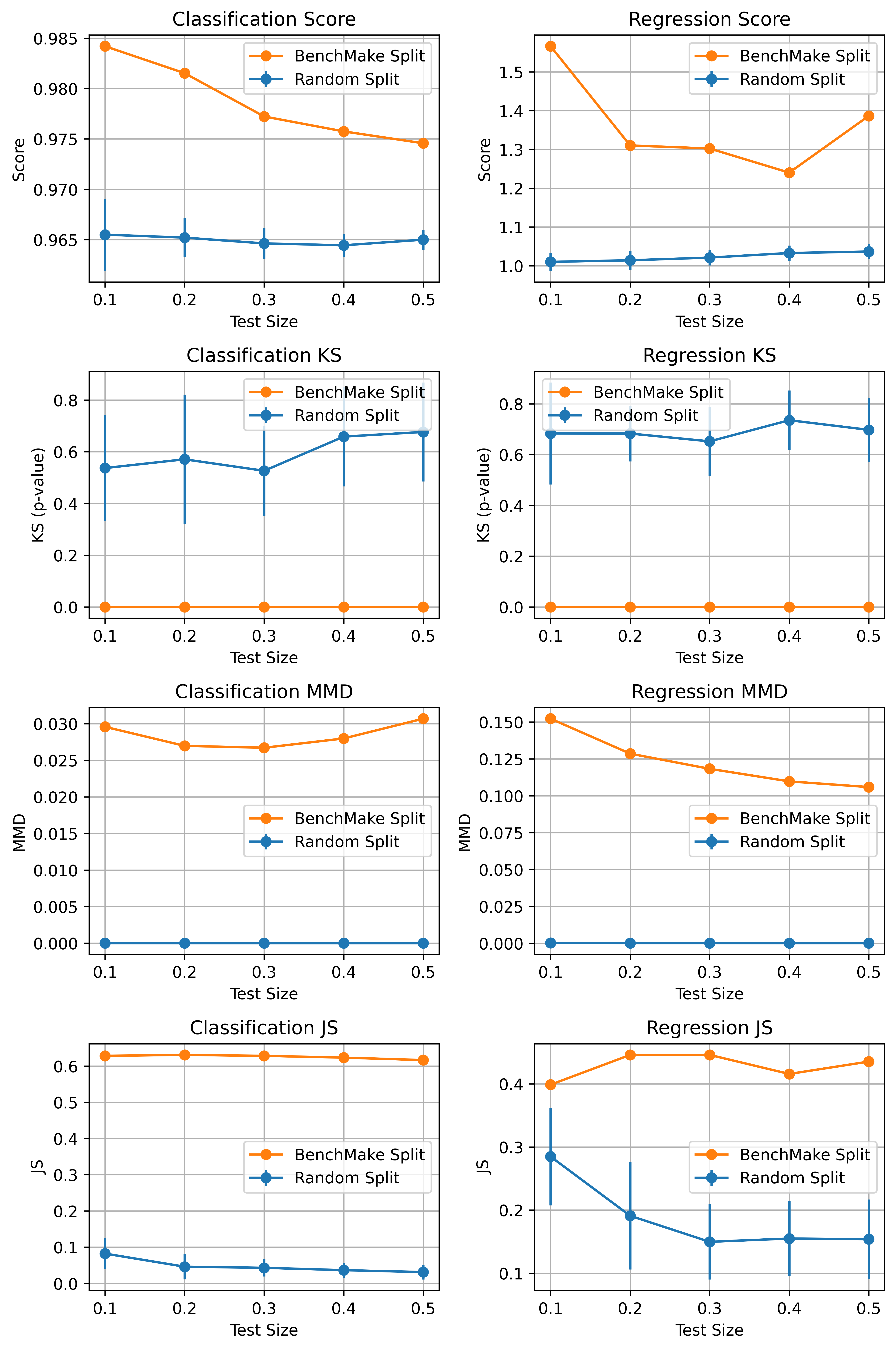}
    \caption{Statistical comparison of the training and testing data for the OGB graph sets, split using random \texttt{train\_test\_split} from scikit-learn (blue) and BenchMake (orange). MOLHIV set for classification is shown on the left, and MOLLIPO for regression is shown on the right.}
    \label{fig:graph_stats}
\end{figure}

In this case the graph structures were numerically encoded for analysis, and the BenchMake split compared to the OGB split and average of 20 random splits; the results are shown in Table~\ref{graphs}. In each case the BenchMake partitions are superior; in some MOLHIV cases by orders of magnitude. This is significant since OGB uses a scaffold split for the data set (random for MOLLIPO). Scaffold splits are generally considered to be better at avoiding data leakage, where molecules in the test set are highly similar to molecules in the training set (sharing the same or very similar scaffolds but have slightly different side chains), by forcing the model to predict properties of molecules with entirely new core structures. Models trained on randomly split data often show inflated performance metrics on the test set, and scaffold splitting is intended to provides a more challenging and realistic evaluation, leading to more reliable estimates of a real-world model performance. The results in Table~\ref{graphs} confirms that the OGB scaffold splits are superior to random splits in some evaluations, but fail to approach the low T-test and KS-test p-values and MI, and high KL or JS divergence and MMD or Wasserstien distance obtained using BenchMake. This suggests that BenchMake splits are likely to offer even higher protection against data leakage and realistic evaluation, without needing to draw on domain knowledge.  Fig.~\ref{fig:graph_umap} {and Fig.~\ref{fig:graph_hist}} compares the train-test distributions, and Fig.~\ref{fig:graph_stats} confirms BenchMake partitioning performance is retained as the size of the testing set increases; BenchMake splits are always more challenging for the classification and regression models.

\subsection{Images}
Image data sets were sourced from MedMNIST~\cite{Yang2020MedMNISTCD}, particularly PneumoniaMNIST for classification tasks and RetinaMNIST for regression. The PneumoniaMNIST data set contains 5,856 pediatric chest X-Ray (grey-scale) images, classifying pneumonia cases against controls with a train-test split of 9:1~\cite{Kermany2018IdentifyingMD}. The images are $(384\,\mathrm{to}\,2,916)\times(127\,\mathrm{to}\,2,713)$, center-cropped with a window size based on the short edge and resized to the MNIST $1\times28\times28$. The RetinaMNIST data set is based on the DeepDRiD24 challenge~\cite{Yang2021MedMNISTV}, containing 1,600 retina fundus images for ordinal regression of the 5-level grading of diabetic retinopathy (DR) severity. The task is to predict a 5-level grade (0 to 4) indicating no DR up to severe DR, as an ordinal regression problem (i.e., 5 classes with an inherent order).  The MedMNIST v2 standard split is stratified using 1080 images for training, 120 for validation and 400 for testing.

For image data, BenchMake expects input in the form of a multi-dimensional array or DataFrame, where each image is typically structured as (\texttt{n\_samples}, \texttt{height}, \texttt{width}, \texttt{channels}). It first converts the data to a float32 NumPy array and then flattens each image (\texttt{n\_samples}, {height}$\times$\texttt{width}$\times$\texttt{channels}) into a one-dimensional vector so that every image is represented as a row vector. The rows are then reordered deterministically using the stable hashing strategy. The images (now as 1D vectors) are Min-Max scaled, and the data is partitioned. BenchMake returns either the training and testing sub-sets in the same format as the original input or the corresponding indices.

\begin{table}[ht]
\centering\small
\caption{Comparison of classification and regression results for data splitting methods for the MedMNIST image data set, representative of medicine. Identical splits sizes are used, and superior results are highlighted in bold.\label{images}}
\begin{tabular}{lccc|ccc}
\hline
\multirow{2}{*}{Metric} & \multicolumn{3}{c|}{Classification (PneumoniaMNIST)} & \multicolumn{3}{c}{Regression (RetinaMNIST)} \\
 & MedMNIST & Random & BenchMake & MedMNIST & Random & BenchMake \\
\hline
T-test (p-value) & 0.1562 & 0.5$\pm$0.1058 & \textbf{0.0000} & 0.2602 & 0.4341$\pm$0.1415 & \textbf{0.0002} \\
KS-test (p-value) & 0.1428 & 0.5$\pm$0.1056 & \textbf{0.0000} & 0.4569 & 0.5799$\pm$0.1450 & \textbf{0.0393} \\
Mutual Information & 0.0021 & 0.0008$\pm$0.0001 & \textbf{0.1042} & 0.0029 & 0.0021 $\pm$ 0.0003 & \textbf{0.1348} \\
KL Divergence & 2.6313 & 1.3$\pm$0.1541 & \textbf{70.5528} & 5.4791 & 2.3658$\pm$0.6402 & \textbf{78.2472} \\
JS Divergence & 0.1889 & 0.088$\pm$0.0042 & \textbf{4.6602} & 0.2353 & 0.1458$\pm$0.0106 & \textbf{3.3362} \\
Wasserstein Distance & 0.0156 & 0.008$\pm$0.0012 & \textbf{0.1870} & 0.0125 & 0.0119$\pm$0.0027 & \textbf{0.1586} \\
Maximum Mean Discrepancy & 0.0019 & 0.0003$\pm$0.0002 & \textbf{0.2182} & 0.0007 & 0.0007$\pm$0.0005 & \textbf{0.1651} \\
\hline
\end{tabular}
\end{table}

\begin{figure}[htbp!] \centering
    \includegraphics[width=1\textwidth]{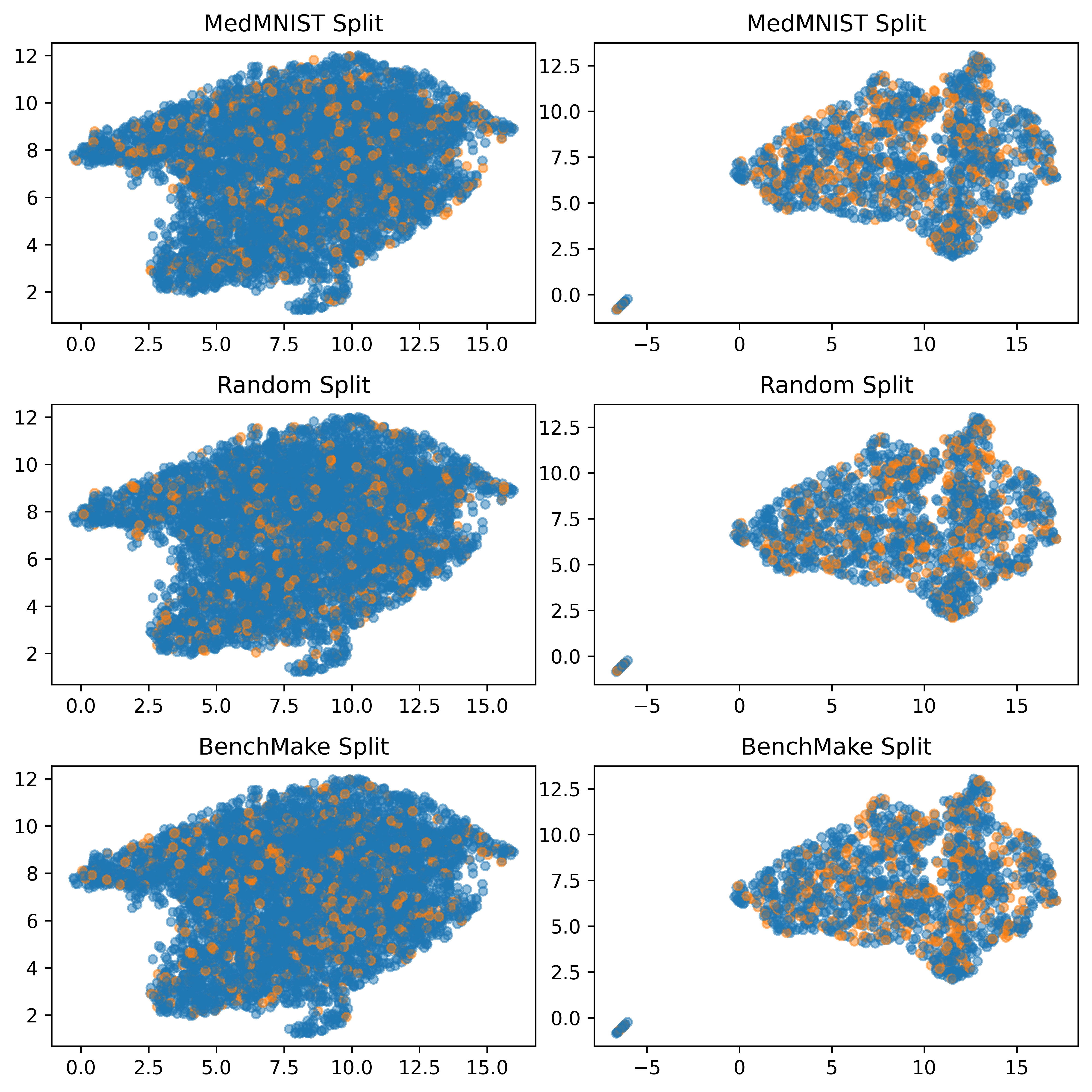}
    \caption{UMAP comparison of the distribution of the training data (blue) and testing data (orange) from the MedMNIST image sets. PneumoniaMNIST set for classification is shown on the left, and RetinaMNIST for regression is shown on the right.}
    \label{fig:image_umap}
\end{figure}

\begin{figure}[htbp!] \centering
    \includegraphics[width=1\textwidth]{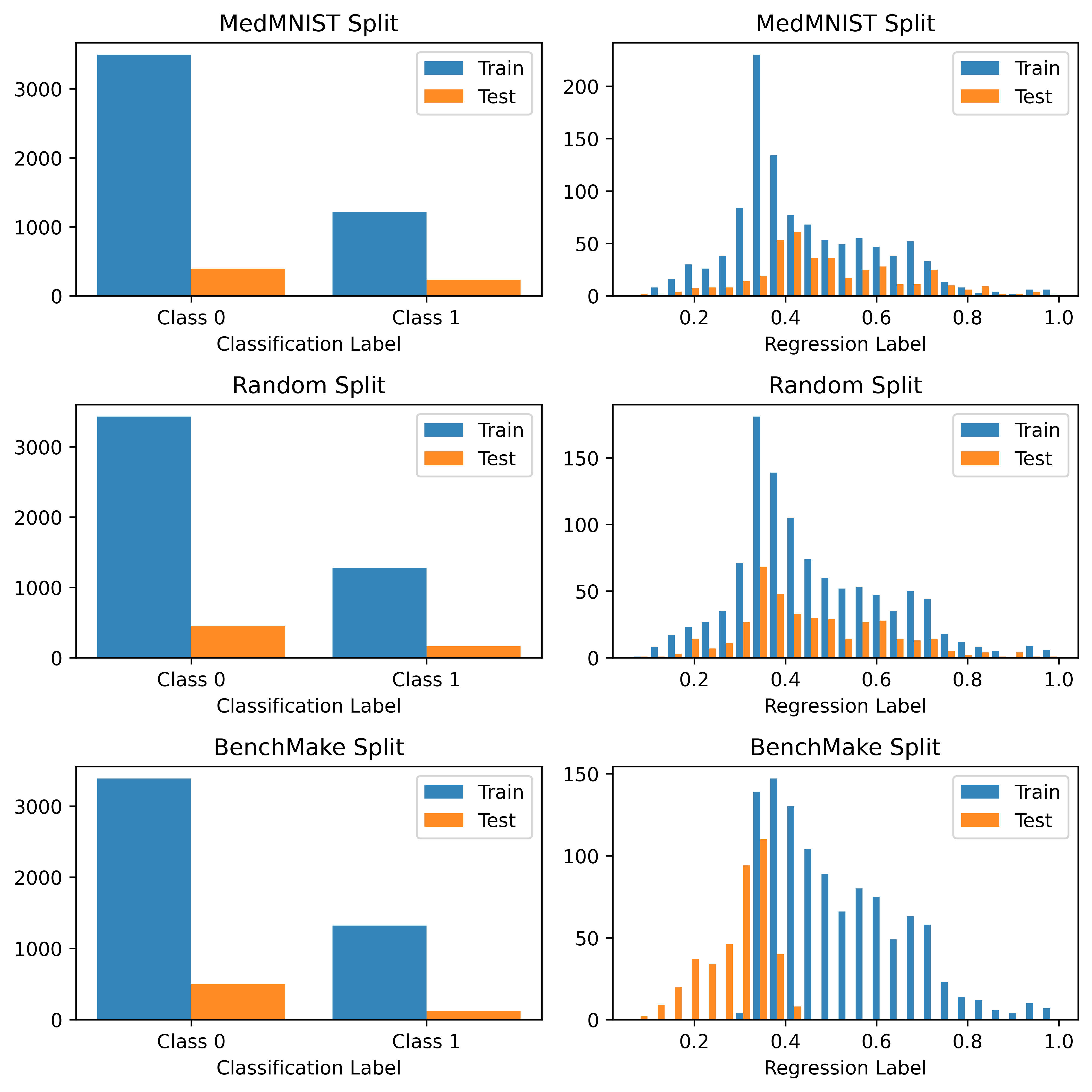}
    \caption{Histogram comparison of the distribution of the training data (blue) and testing data (orange) from the MedMNIST image sets. PneumoniaMNIST set for classification is shown on the left, and RetinaMNIST for regression is shown on the right.}
    \label{fig:image_hist}
\end{figure}

\begin{figure}[htbp!] \centering
    \includegraphics[width=0.8\textwidth]{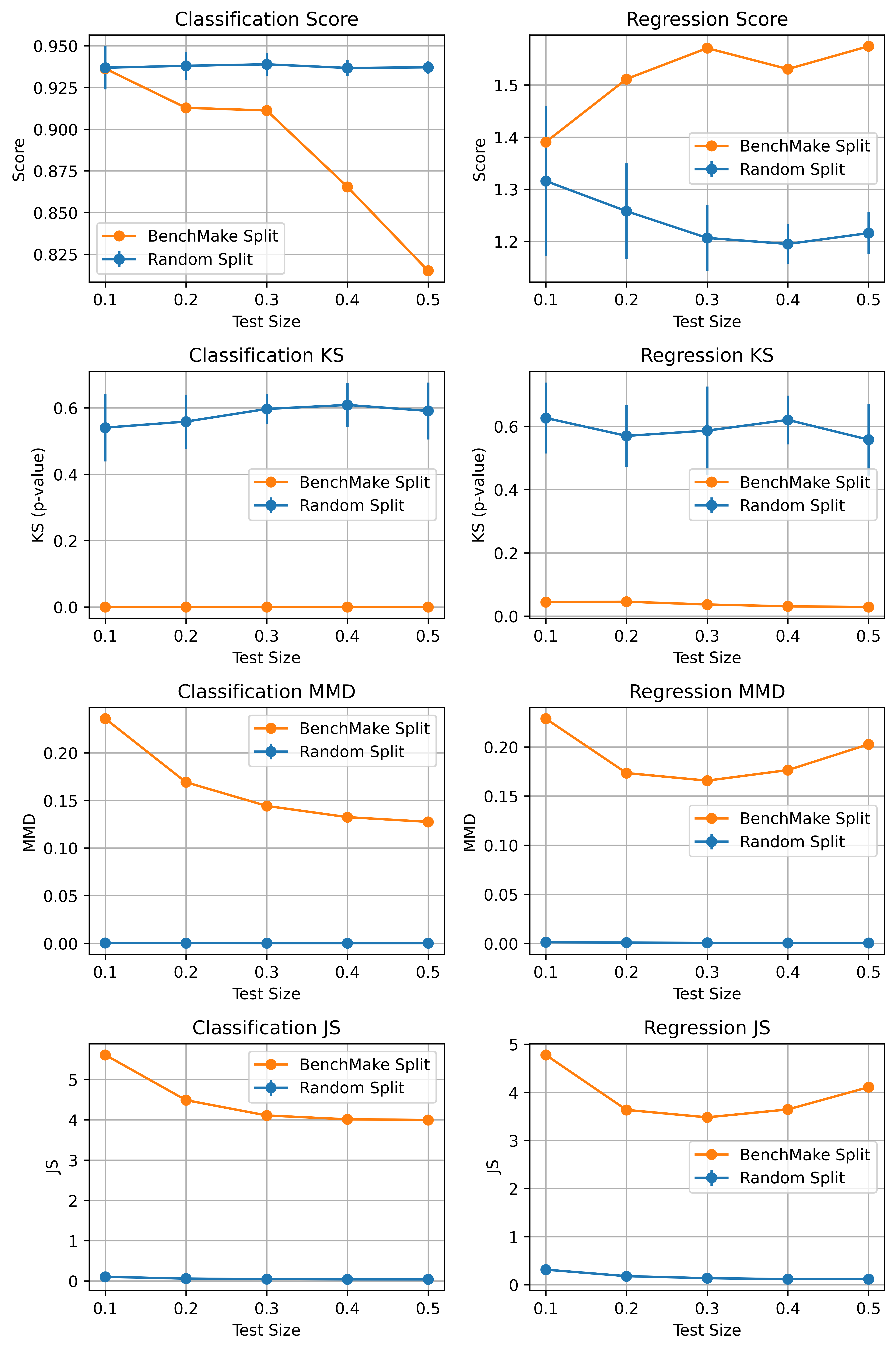}
    \caption{Statistical comparison of the training and testing data for the MedMNIST image sets, split using random \texttt{train\_test\_split} from scikit-learn (blue) and BenchMake (orange). PneumoniaMNIST set for classification is shown on the left, and RetinaMNIST for regression is shown on the right.}
    \label{fig:image_stats}
\end{figure}

Table~\ref{images} contains the results comparing MedMNIST, random and BenchMake splits for classification (PneumoniaMNIST) and regression (RetinaMNIST) data sets.  The MedMNIST splits are almost always superior to the random splits in these seven statistical tests, but still at least one order or magnitude inferior to the BenchMake partitioning. A stand-out result is the KL divergence, where the values are high across all splits, but extremely high for the BenchMake partition. BenchMake automatically aligns the histograms so this discrepancy is valid. This example uses unsupervised learning and certain feature ranges are characteristics of the test set, set while entirely different ranges are characteristic of the training set, resulting in nearly disjoint sets with minimal overlap. This is not immediately apparent in the visual inspection of the UMAPs in Fig.~\ref{fig:image_umap} {and the histograms in Fig.~\ref{fig:image_hist}}, but contributes to the lower model performance but higher statistical performance as the test set size increases (Fig.~\ref{fig:image_stats}). {The regression results (right column in Fig.~\ref{fig:image_hist}) show that the testing set is not evenly distributed across the label space; this is not an error.
This is common when (a) the data is highly unstructured, (b) the relationship between features and target is not strictly monotonic, (c) the data is noisy, and/or (d) the splitting algorithm (such as BenchMake) operates only on the feature space. This protects against data leakage since it is guaranteed to produce test sets that do not represent the label, and provides important information about the data set, highlighting a relationship between the edge cases and the target label.  This relationship has not been identified in previous studies using this set.}

\subsection{Sequential Strings}
The biophysical BACE-1 data set is a collection of 1,513 molecules with activity against human $\beta$-secretase 1 (BACE-1), a drug target relevant in Alzheimer’s disease~\cite{Subramanian2016ComputationalMO}. Released in MoleculeNet~\cite{Wu2017MoleculeNetAB}, it includes both quantitative (IC50 values) and qualitative (binary labels) binding results, specifically split into active (IC50 $\leq$ 100 nM) and inactive classes. MoleculeNet implements ECFP (Extended-Connectivity Fingerprints) featurisation to decompose molecules into sub-modules from heavy atoms, with an assigned unique identifier~\cite{Quirs2018UsingSS}. MoleculeNet recommend scaffold splitting, and the DeepChem~\cite{deepchem} loader default scaffold split is uses in this study.  QM7 is a quantum chemistry data set of small organic molecules for regression tasks, computed by quantum chemical methods~\cite{Rupp2011FastAA}, and distributed by MoleculeNet. The data set contains 7,165 molecules with up to 7 heavy atoms (C, N, O, S) from the GDB-13 chemical space, separated using a stratified split of molecules sorted by energy to provide a uniform sampling (to avoid energy-range bias). The goal is to predict the molecular atomization energy (in kcal/mol) for each molecule.

BenchMake handles sequential data such as text strings (e.g. SMILES strings or DNA sequences) by taking a list or Pandas Series and converting each sequence into a numerical (vector) representation using a character-level \texttt{CountVectorizer}. This transformation results in a two-dimensional NumPy array (float32) where each row corresponds to the numeric representation of a sequence, which are deterministically reordered via the stable hash.  BenchMake then applies Min-Max scaling and partitions the data. The original sequences are then re-ordered using the same hash order, and BenchMake returns either the full training and testing splits (list or series) or the indices of the splits.

\begin{table}[ht]
\centering\small
\caption{Comparison of classification and regression results for data splitting methods for the MoleculeNet sequential strings, representative of chemistry. Identical splits sizes are used, and superior results are highlighted in bold.\label{sequences}}
\begin{tabular}{lccc|ccc}
\hline
\multirow{2}{*}{Metric} & \multicolumn{3}{c|}{Classification (BACE-1)} & \multicolumn{3}{c}{Regression (QM7)} \\
 & Scaffold & Random & BenchMake & Uniform & Random & BenchMake \\
\hline
T-test (p-value)   & \textbf{0.5336} & 0.54$\pm$0.0174 & 0.5388 & 0.5188 & 0.518$\pm$0.0057 & \textbf{0.5148} \\
KS-test (p-value)   & \textbf{0.9928} & 0.994$\pm$0.0047 & 0.9962 & 0.9990 & 0.9990$\pm$0.0007 & 0.9990 \\
Mutual Information & 0.0016 & 0.0016$\pm$0.0001 & 0.0016 & 0.0007 & 0.0007$\pm$0.0001 & 0.0007 \\
KL Divergence   & 0.0017 & \textbf{0.0018$\pm$0.0003} & 0.0016 & 0.0004 & 0.0004$\pm$0.0001 & 0.0004 \\
JS Divergence    & 0.0004 & 0.0004$\pm$0.0001 & 0.0004 & 0.0001 & 0.0001$\pm$0.0001 & 0.0001 \\
Wasserstein Distance & \textbf{0.0074} & 0.0071$\pm$0.0007 & 0.0067 & 0.0016 & 0.0016$\pm$0.0001 & 0.0016 \\
Maximum Mean Discrepancy  & \textbf{0.0085} & 0.008$\pm$0.0008 & 0.0076 & 0.0027 & 0.0027$\pm$0.0001 & 0.0027 \\
\hline
\end{tabular}
\end{table}

\begin{figure}[htbp!] \centering
    \includegraphics[width=1\textwidth]{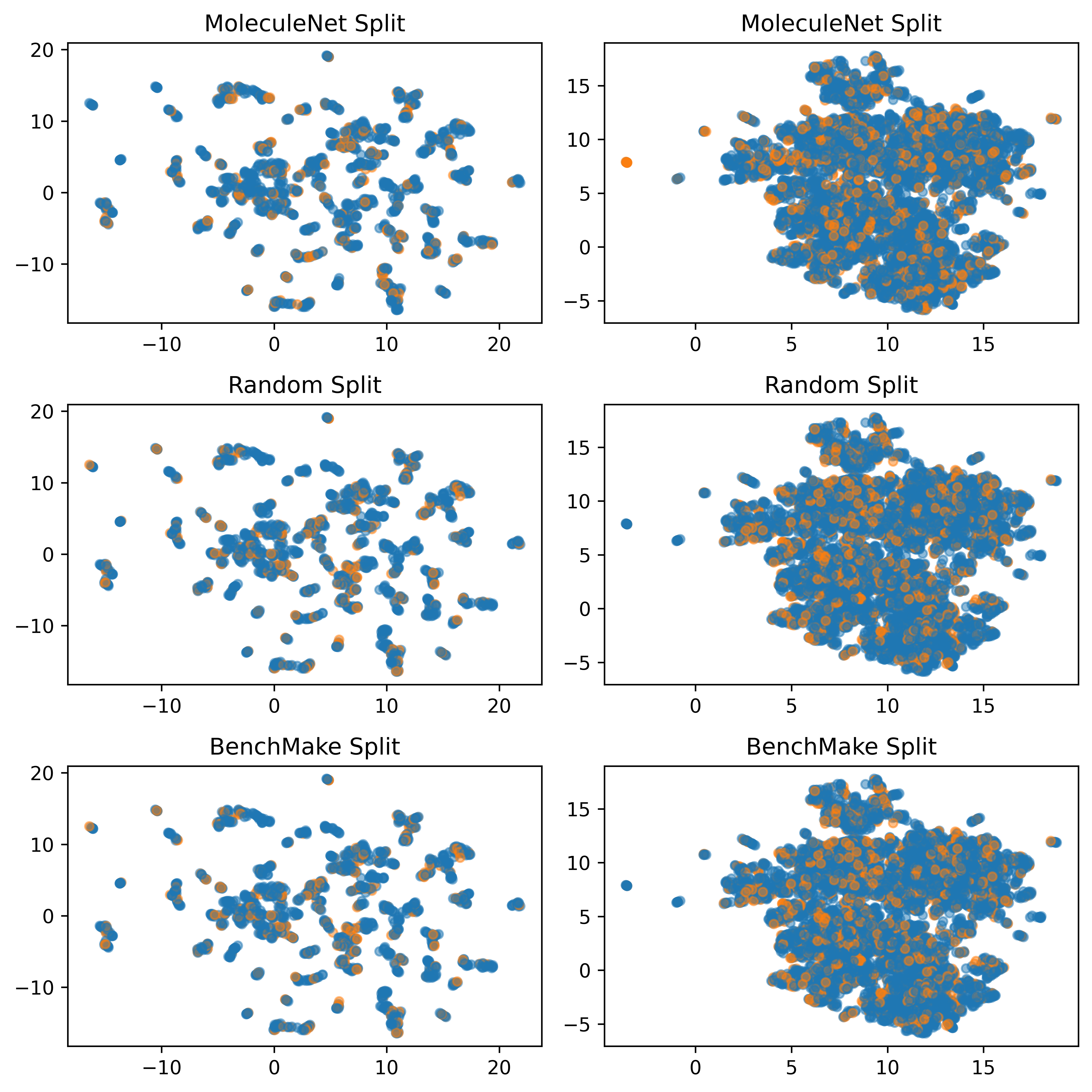}
    \caption{UMAP comparison of the distribution of the training data (blue) and testing data (orange) from the MoleculeNet sequential strings (text). BACE-1 set for classification is shown on the left, and QM7 for regression is shown on the right.}
    \label{fig:sequence_umap}
\end{figure}

\begin{figure}[htbp!] \centering
    \includegraphics[width=1\textwidth]{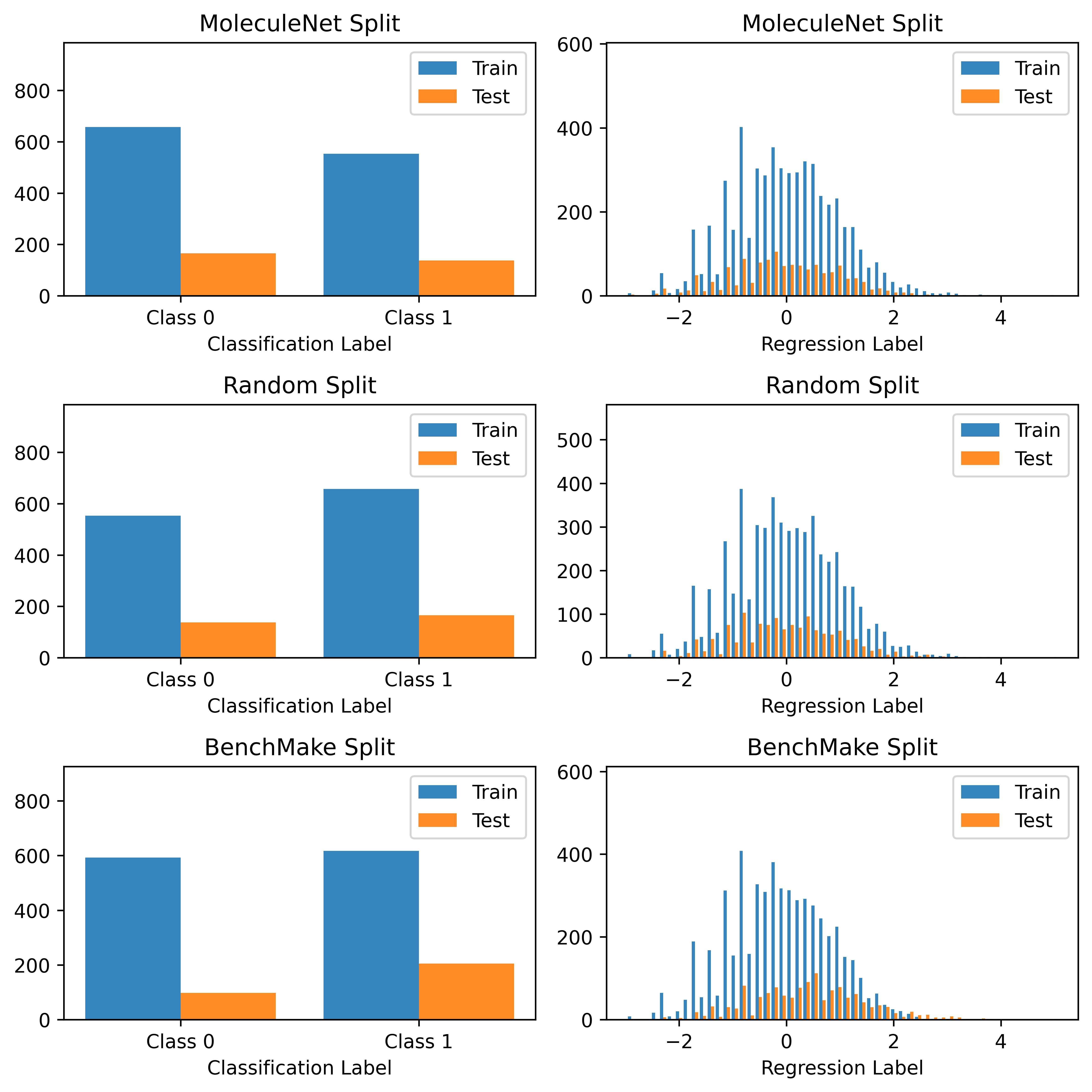}
    \caption{Histogram comparison of the distribution of the training data (blue) and testing data (orange) from the MoleculeNet sequential strings (text). BACE-1 set for classification is shown on the left, and QM7 for regression is shown on the right.}
    \label{fig:sequence_hist}
\end{figure}

\begin{figure}[htbp!] \centering
    \includegraphics[width=0.8\textwidth]{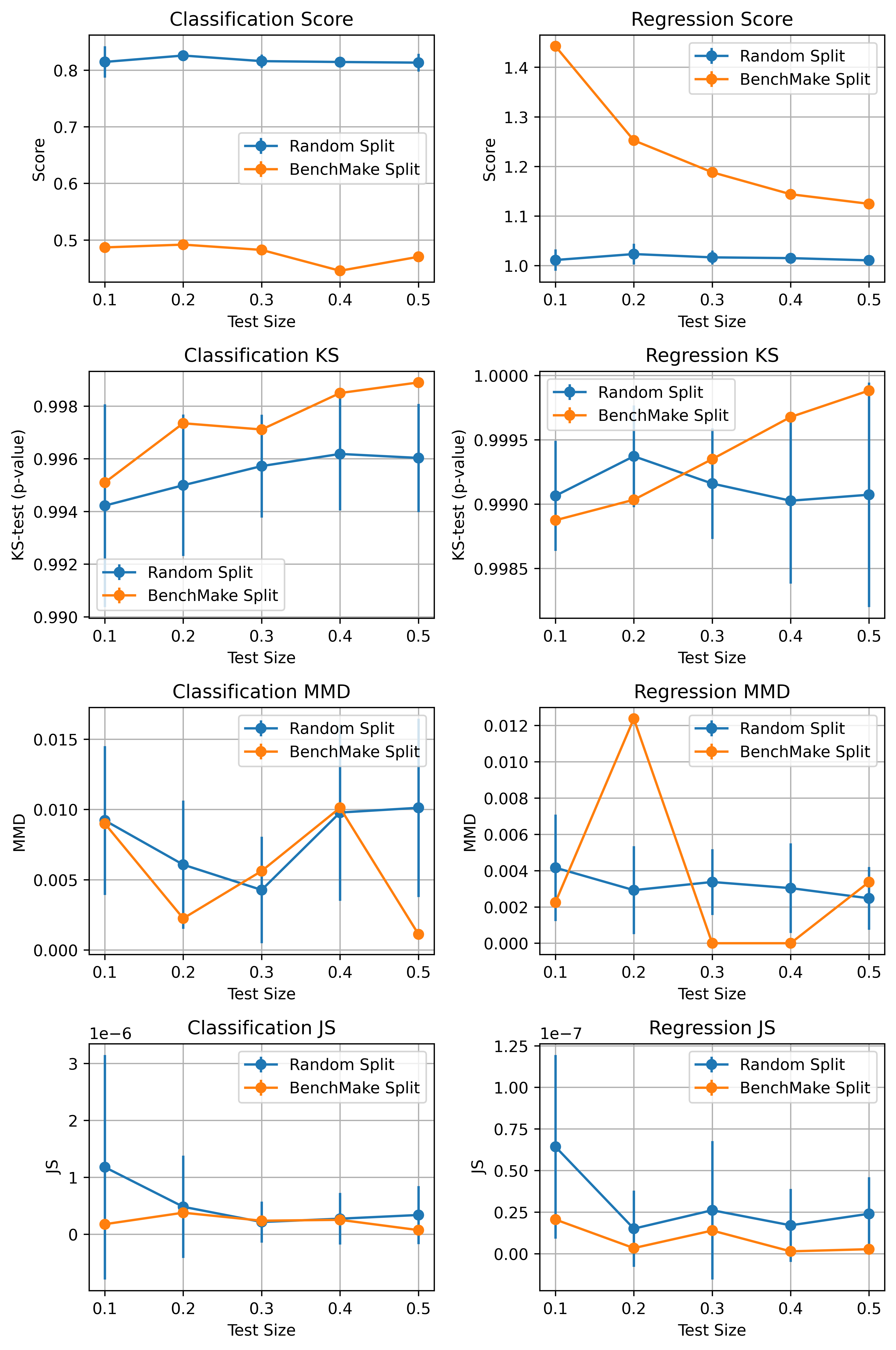}
    \caption{Statistical comparison of the training and testing data for the MoleculeNet sequential string (text) sets, split using random \texttt{train\_test\_split} from scikit-learn (blue) and BenchMake (orange). BACE-1 set for classification is shown on the left, and QM7 for regression is shown on the right.}
    \label{fig:sequence_stats}
\end{figure}

In this example, the splits compared in Table~\ref{sequences} are virtually indistinguishable, even though scaffold splitting is chemically-enriched and the preferred approach in this domain.  While the SMILES strings in the BACE-1 and QM7 data sets are text, these collections do not exhibit a meaningful sequential order (such as time or document sequence) that would create a distribution shift. As a result, similar sub-sets are obtained regardless of whether they are split randomly or partitioned sequentially with BenchMake, and the statistical metrics are the same. Under these circumstances, BenchMake performs as expected and is consistent with the scikit-learn \texttt{train\_test\_split}; no better or worse than random.  Distributions are compared in Fig.~\ref{fig:sequence_umap} {and Fig.~\ref{fig:sequence_hist}}.  Regardless of the statistical evaluations, the model performances in Fig.~\ref{fig:sequence_stats} do highlight that the BenchMake partitions are consistently more challenging.

\subsection{Signals}
The ST003210 (Single-cell Metabolomics of TNBC Cells) is part of the Metabolomics Workbench repository~\cite{metabolomics} (Project ID PR002001). This data set contains untargeted metabolomics data from single cells and cell clusters of two triple-negative breast cancer (TNBC) cell lines.  The original goal was to explore metabolic differences between individual cancer cells and clusters of cells, relating to metastatic potential, and the task here is a binary classification (single cell vs cluster) based on metabolomic profile. The data set includes metabolomic profiles for 80 samples of cell extracts, with 4 groups: single cells from cell line MDA-MB-231, cell clusters from MDA-MB-231, single cells from MDA-MB-453, cell clusters from MDA-MB-453 (approximately 10 biological replicates per group, plus some QC and blanks).  Blanks and QCs were omitted from this example which classified single cells versus cell clusters, and a random train-test split of 0.2 was used since no predefined split was available from the Metabolomics Workbench.

The Greenhouse Gas Observing Network (GGON) data set provides time‐series measurements from sensors that continuously record the concentrations of key greenhouse gases from a network of 2921 grid cells in California ($12\times12$ km$^2$), created using simulations of the Weather Research and Forecast model with Chemistry (WRF-Chem) over May 10 to July 31, 2010~\cite{ghg}. Each file in the data set represents a signal data from one site, containing 16 times series of green house gas (GHG) emissions spaced 6 hours apart, with the site identifier extracted from the filename used as the target variable for regression. These stations are part of an observing network that systematically collects data over time, providing a multi-dimensional view of how atmospheric greenhouse gas levels vary.  All GHG concentrations are in units of parts per trillion, and a random train-test split of 0.2 was used since no predefined split was available from the UCI Machine Learning Repository~\cite{ghg}.

For signal data such as spectra, time series, audio signals, or sensor outputs BenchMake first ensures that the data is represented as a consistent float32 NumPy array. If the signals are provided in a multi-dimensional format (for example, if each signal has multiple channels or time-points arranged in a 3D array), they are flattened so that each signal becomes a single row vector. Once in this unified 2D format, the rows are deterministically sorted using stable hashing method, and then standardised with Min-Max scaling prior to partitioning.  BenchMake returns either the resulting training and testing data in the same structure as the input (e.g., NumPy arrays or DataFrames) or simply the lists of indices for each split.

\begin{table}[ht]
\centering\small
\caption{Comparison of Classification and Regression Results for data splitting methods for the open signal data sets representative of scientific spectra and time series.  Identical splits sizes are used, and superior results are highlighted in bold.\label{signals}}
\begin{tabular}{lccc|ccc}
\hline
\multirow{2}{*}{Metric} & \multicolumn{3}{c|}{Classification (ST003210)} & \multicolumn{3}{c}{Regression (GGON)} \\
 & Manual & Random & BenchMake & Manual & Random & BenchMake \\
\hline
T-test (p-value) & 0.4818 & 0.5$\pm$0.1100 & \textbf{0.2070 }& 0.4865 & 0.50$\pm$0.0395 & \textbf{0.0505 }\\
KS-test (p-value) & 0.6980 & 0.7$\pm$0.1018 & \textbf{0.3079} & 0.3819 & 0.57$\pm$0.0733 & \textbf{0.0007} \\
Mutual Information & 0.0055 & 0.0055$\pm$0.0004 & \textbf{0.0264} & 0.0010 & 0.0010$\pm$0.0001 & \textbf{0.0516 }\\
KL Divergence & 0.1165 & 0.08$\pm$0.0251 &\textbf{1.0778} & 0.0049 & 0.004$\pm$0.0022 & \textbf{1.8056 }\\
JS Divergence & 0.0280 & 0.020$\pm$0.0062 & \textbf{0.2280} & 0.0012 & 0.0009$\pm$0.0006 & \textbf{0.3317}\\
Wasserstein Distance  & 0.0074 & 0.009$\pm$0.0068 & \textbf{0.0305} & 0.0006 & 0.0007$\pm$0.0003 & \textbf{0.0165 }\\
Maximum Mean Discrepancy   & 0.0010 & 0.005$\pm$0.0030 & \textbf{0.0425}& 0.0013 & 0.003$\pm$0.0019 & \textbf{0.0195 }\\
\hline
\end{tabular}
\end{table}

\begin{figure}[htbp!] \centering
    \includegraphics[width=1\textwidth]{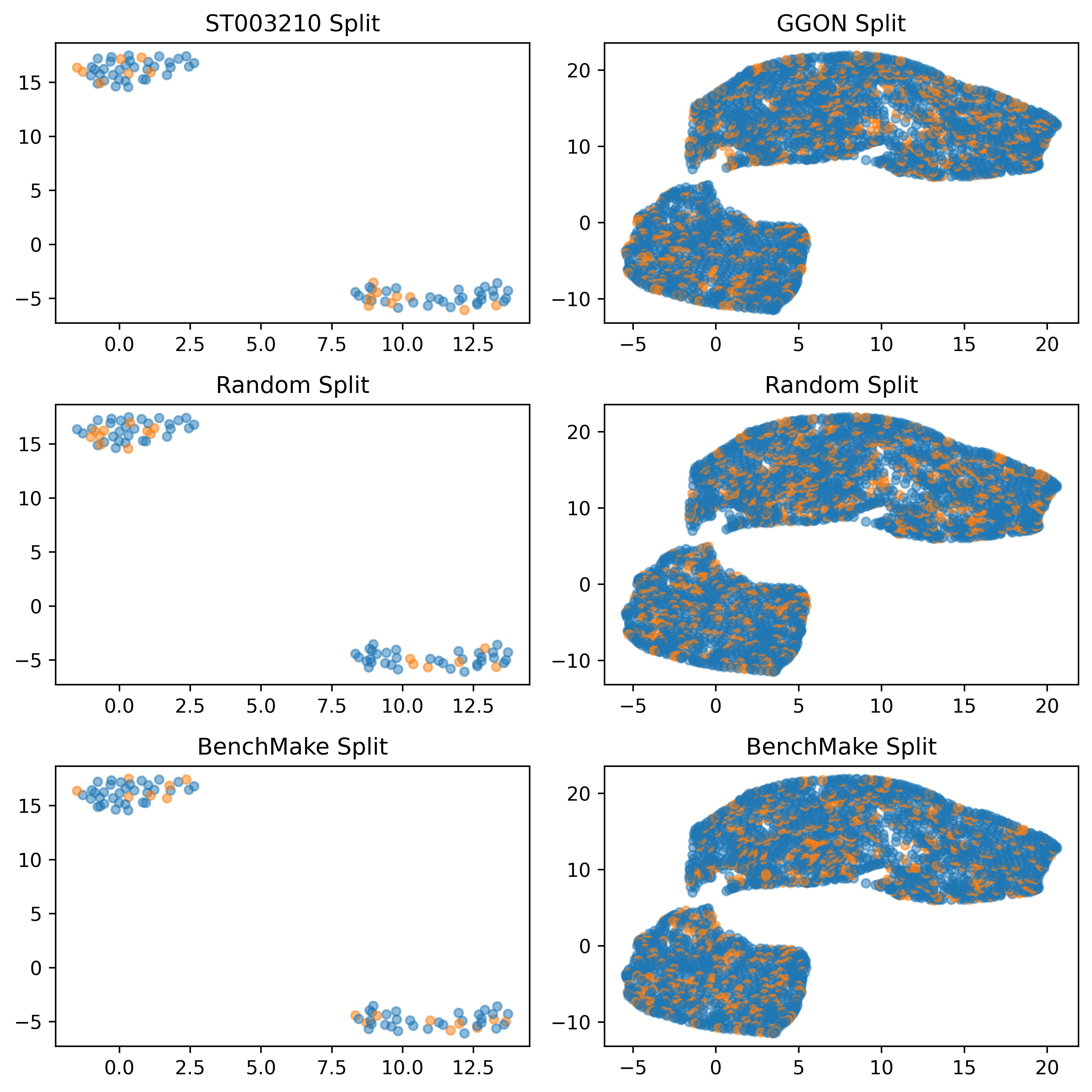}
    \caption{UMAP comparison of the distribution of the training data (blue) and testing data (orange) from open scientific spectra and signal sets. ST003210 set from the Metabolomics Workbench for classification is shown on the left, and the GGON from UCI Machine Learning Repository for regression is shown on the right.}
    \label{fig:signal_umap}
\end{figure}

\begin{figure}[htbp!] \centering
    \includegraphics[width=1\textwidth]{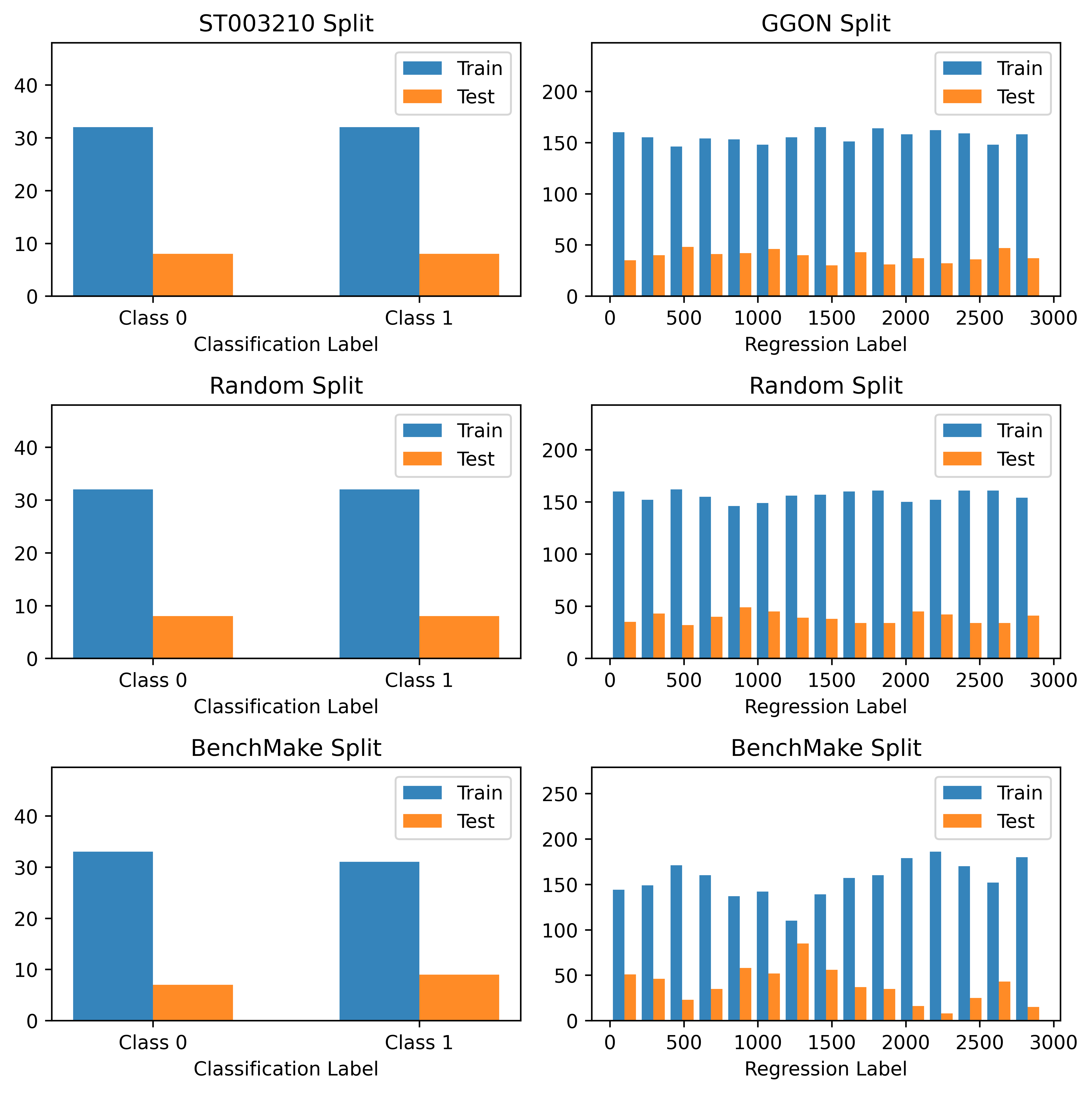}
    \caption{Histogram comparison of the distribution of the training data (blue) and testing data (orange) from open scientific spectra and signal sets. ST003210 set from the Metabolomics Workbench for classification is shown on the left, and the GGON from UCI Machine Learning Repository for regression is shown on the right.}
    \label{fig:signal_hist}
\end{figure}

\begin{figure}[htbp!] \centering
    \includegraphics[width=0.8\textwidth]{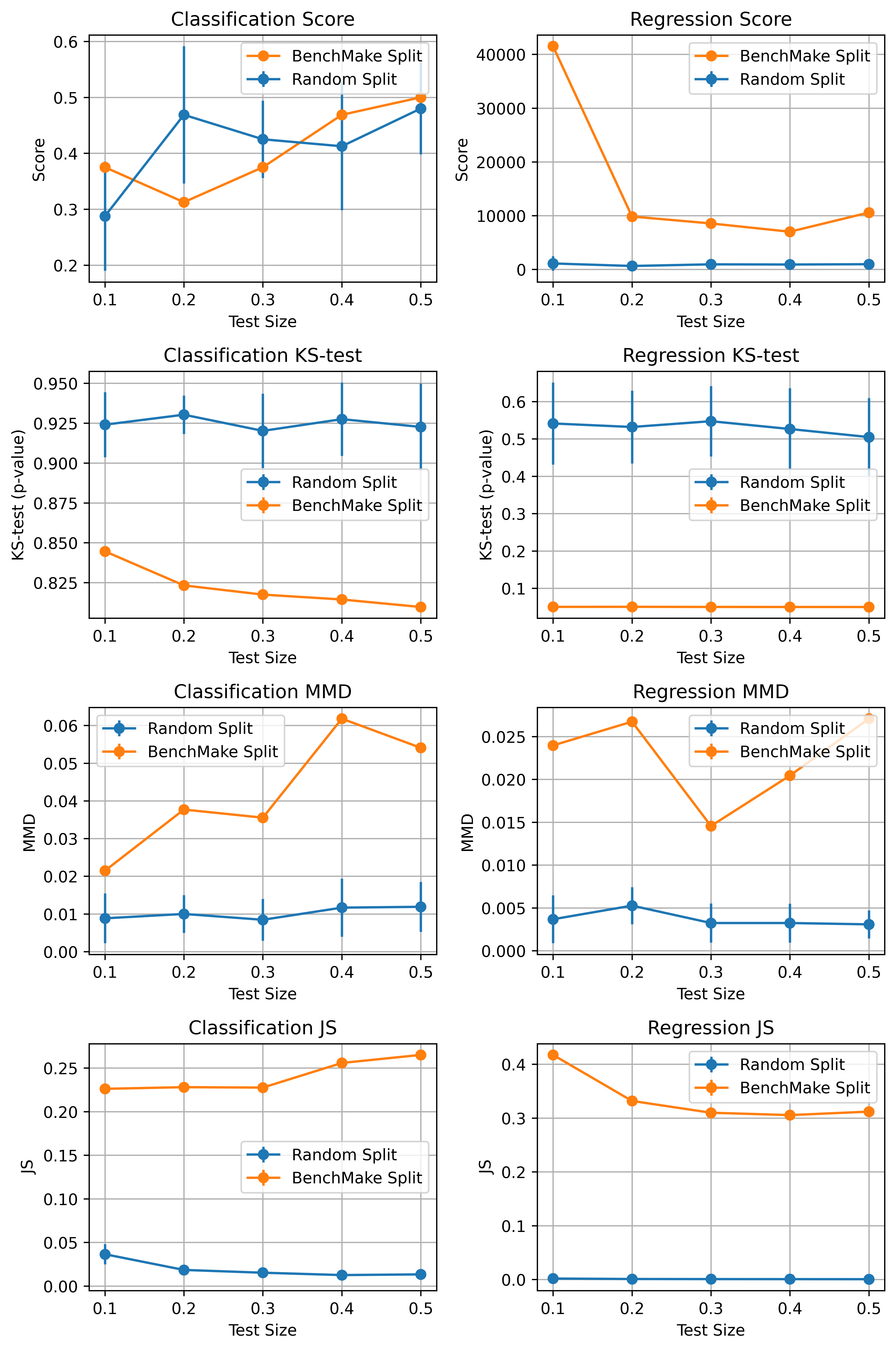}
    \caption{Statistical comparison of the training and testing data for the open scientific spectra and signal sets, split using random \texttt{train\_test\_split} from scikit-learn (blue) and BenchMake (orange). ST003210 set from the Metabolomics Workbench for classification is shown on the left, and the GGON from UCI Machine Learning Repository for regression is shown on the right.}
    \label{fig:signal_stats}
\end{figure}

Evaluation of the splitting of the signal data sets is shown in Table~\ref{signals}, where, once again, BenchMake outperforms the random sampling, partitioning testing sets with more diverse distributions, and with greater divergence and discrepancy (compared to the training sets). In the case of ST003210 classification, the difference with the BenchMake partition increases with increasing test set size (Fig.~\ref{fig:signal_stats}), even though the model performance is similar to the random splits. Regression is always more challenging using the BenchMake partitions. The {distributions} in Fig.~\ref{fig:signal_umap} {and Fig.~\ref{fig:signal_hist}} highlight that the GGON data set exhibits a binary distribution conducive to classification.

\section{Conclusion}

More widespread benchmarking of machine learning models using appropriate scientific data sets will help to progress many fields in computational science, providing reproducible evidence of model performance, stability, efficiency and trust. The extreme diversity of science problems, and the complexity and variability of the data sets they produce, make it difficult to identify appropriate benchmarks, even as open access scientific data sets are becoming more plentiful.  BenchMake is a convenient and versatile pip installable Python package that can ingest data in a variety of formats and partition the most challenging edge cases into a testing set of any size; deterministically turning any scientific data set into a benchmark.

By identifying and matching archetypes with data points, BenchMake can partition more meaningful test cases, especially in domains where diversity matters, such as chemical compounds and image classification. This challenges models to perform better on the extremes of a data set, supporting better generalisation. BenchMake has been evaluated on 10 different published benchmark data sets, including both classification and regression tasks with 5 different data formats, and in almost all cases outperforms the prescribed train-test and random splits, across 7 statistical tests. The data sets represented 6 different scientific disciplines, with a variety of sizes, dimensionalities and distributions. {All 5 classification sets  used to demonstrate the capabilities are imbalanced.}  Model scores are shown to be lower for 8 of the 10 BenchMake testing sets, confirming that the testing sets are more challenging.  Ideal for machine learning challenges and competitions, researchers can simply report the data set and test set fraction used, and others can use BenchMake to benchmark under the same conditions. {Since BenchMake is entirely unsupervised, using non-negative matrix factorisation on the feature space, it avoids data leakage (a risk in domain-splitting methods such as scaffold splits or active learning-based approaches that use the training and testing set during sampling~\cite{Dong2024OnlineMG}), is reproducible (unlike random methods), and maximises the difference between the sets (unlike CoreSets~\cite{coresets}). However, BenchMake is static and must be re-run if new data is added to the set.}

A currently limitation of BenchMake is the computational intensity, scaling as $\mathcal{O}(n^2\times d)$, where $n$ is the number of data instances and $d$ is the dimensionality, and the overall complexity depends on the number of iterations needed for convergence. While BenchMake is GPU enabled and uses parallel processing, not all parts of the code are parallelisable, and the development or introduction of more advanced implementations of NMF may improve efficiency.  Currently, the slower partitioning time (relative to random splitting) should be traded-off against the reproducibility and superior test set characteristics. Although there are existing tools for train-test splitting, resampling and feature extraction which provide representative sub-sets, BenchMake explicitly isolates challenging, archetypal edge cases, potentially exposing model limitations and biases that conventional methods might overlook.

\section{Acknowledgments}
Computational resources for this project were supplied by the National Computing Infrastructure (NCI) [grant number p00]. A. S. Barnard would like to thank Chloe Lin, Amanda Parker, Ben Mashford and Dan Andrews for beta testing.

\section{Conflict of interest}
There is no conflict of interest to declare.

\section{Data availability statement}
No data was generated as part of this work. BenchMake is available at \url{https://pypi.org/project/benchmake/}.

\clearpage
\section{Appendix A --- Supporting Information}

\pagebreak
\bibliographystyle{unsrt}
\bibliography{references}

\begin{thebibliography}{10}

\bibitem{oliver2018realistic}
Avital Oliver, Augustus Odena, Colin~A Raffel, Ekin~Dogus Cubuk, and Ian
  Goodfellow.
\newblock Realistic evaluation of deep semi-supervised learning algorithms.
\newblock {\em Advances in neural information processing systems}, 31, 2018.

\bibitem{deng2009imagenet}
Jia Deng, Wei Dong, Richard Socher, Li-Jia Li, Kai Li, and Li~Fei-Fei.
\newblock Imagenet: A large-scale hierarchical image database.
\newblock In {\em 2009 IEEE Conference on Computer Vision and Pattern
  Recognition}, pages 248--255. IEEE, 2009.

\bibitem{lecun2015deep}
Yann LeCun, Yoshua Bengio, and Geoffrey Hinton.
\newblock Deep learning.
\newblock {\em Nature}, 521(7553):436--444, 2015.

\bibitem{jordan2015machine}
Michael~I Jordan and Tom~M Mitchell.
\newblock Machine learning: Trends, perspectives, and prospects.
\newblock {\em Science}, 349(6245):255--260, 2015.

\bibitem{zhang2021understanding}
Chiyuan Zhang, Samy Bengio, Moritz Hardt, Benjamin Recht, and Oriol Vinyals.
\newblock Understanding deep learning requires rethinking generalization.
\newblock {\em Communications of the ACM}, 64(3):107--115, 2021.

\bibitem{krizhevsky2009learning}
Alex Krizhevsky and Geoffrey Hinton.
\newblock Learning multiple layers of features from tiny images.
\newblock Technical report, University of Toronto, 2009.

\bibitem{vanschoren2014openml}
Joaquin Vanschoren, Jan~N Van~Rijn, Bernd Bischl, and Luis Torgo.
\newblock Openml: networked science in machine learning.
\newblock {\em ACM SIGKDD Explorations Newsletter}, 15(2):49--60, 2014.

\bibitem{Mahmood2025ABC}
Faisal Mahmood.
\newblock A benchmarking crisis in biomedical machine learning.
\newblock {\em Nature Medicine}, 31:1060, 2025.

\bibitem{gil2016toward}
Yolanda Gil et~al.
\newblock Toward the geoscience paper of the future: Best practices for
  documenting and sharing research from data to software to provenance.
\newblock {\em Earth and Space Science}, 3:388--415, 2016.

\bibitem{wagstaff2012machine}
Kiri Wagstaff.
\newblock Machine learning that matters.
\newblock {\em arXiv preprint arXiv:1206.4656}, 2012.

\bibitem{bossuyt2003stard}
Patrick~M Bossuyt, Johannes~B Reitsma, David~E Bruns, Constantine~A Gatsonis,
  et~al.
\newblock The stard statement for reporting studies of diagnostic accuracy:
  explanation and elaboration.
\newblock {\em Clinical Chemistry}, 49(1):7--18, 2003.

\bibitem{weber2019uncertainty}
T~Weber and P~Bowyer.
\newblock Uncertainty in climate science: a multi-model approach.
\newblock {\em Earth System Dynamics}, 10(4):701--716, 2019.

\bibitem{ramakrishnan2014quantum}
Raghunathan Ramakrishnan, Pavlo~O Dral, Matthias Rupp, and O~Anatole von
  Lilienfeld.
\newblock Quantum chemistry structures and properties of 134 kilo molecules.
\newblock {\em Scientific Data}, 1:140022, 2014.

\bibitem{street1993nuclear}
W~Nick Street, William~H Wolberg, and Olvi~L Mangasarian.
\newblock Nuclear feature extraction for breast tumor diagnosis.
\newblock {\em IS\&T/SPIE International Symposium on Electronic Imaging:
  Science and Technology}, pages 861--870, 1993.

\bibitem{wang2017chestx}
Xiaosong Wang, Yifan Peng, Le~Lu, Zhiyong Lu, Mohammadhadi Bagheri, and
  Ronald~M. Summers.
\newblock Chestx-ray: Hospital-scale chest x-ray database and benchmarks on
  weakly supervised classification and localization of common thorax diseases.
\newblock pages 369--392, 2019.

\bibitem{pasquet2019ethical}
Adrien Pasquet et~al.
\newblock The ethical implications of big data research in human biology.
\newblock {\em BMC Medical Ethics}, 20(1):1--12, 2019.

\bibitem{lecun1998gradient}
Yann LeCun, Leon Bottou, Yoshua Bengio, and Patrick Haffner.
\newblock Gradient-based learning applied to document recognition.
\newblock {\em Proceedings of the IEEE}, 86(11):2278--2324, 1998.

\bibitem{harrison1978hedonic}
David Harrison and Daniel~L Rubinfeld.
\newblock Hedonic housing prices and the demand for clean air.
\newblock {\em Journal of Environmental Economics and Management},
  5(1):81--102, 1978.

\bibitem{liu2015deep}
Ziwei Liu, Ping Luo, Xiaogang Wang, and Xiaoou Tang.
\newblock Deep learning face attributes in the wild.
\newblock In {\em Proceedings of the IEEE International Conference on Computer
  Vision}, pages 3730--3738, 2015.

\bibitem{Javadi2017NonnegativeMF}
Hamid Haj~Seyyed Javadi and Andrea Montanari.
\newblock Nonnegative matrix factorization via archetypal analysis.
\newblock {\em Journal of the American Statistical Association}, 115:896 --
  907, 2017.

\bibitem{edge}
Niklas Bunzel, Nicolas G\"{o}ller, and Raphael~Antonius Frick.
\newblock Identifying and generating edge cases.
\newblock In {\em Proceedings of the 2nd ACM Workshop on Secure and Trustworthy
  Deep Learning Systems}, SecTL '24, page 16–23, New York, NY, USA, 2024.
  Association for Computing Machinery.

\bibitem{Barber1996TheQA}
C.~Bradford Barber, David~P. Dobkin, and Hannu Huhdanpaa.
\newblock The quickhull algorithm for convex hulls.
\newblock {\em ACM Transactions on Mathematical Software (TOMS)}, 22:469 --
  483, 1996.

\bibitem{Lee1999LearningTP}
Daniel~D. Lee and H.~Sebastian Seung.
\newblock Learning the parts of objects by non-negative matrix factorization.
\newblock {\em Nature}, 401:788--791, 1999.

\bibitem{cutler1994archetypal}
Adele Cutler and Leo Breiman.
\newblock Archetypal analysis.
\newblock {\em Technometrics}, 36(4):338--347, 1994.

\bibitem{Mrup2010ArchetypalAF}
Morten M{\o}rup and Lars~Kai Hansen.
\newblock Archetypal analysis for machine learning.
\newblock {\em IEEE International Workshop on Machine Learning for Signal
  Processing}, pages 172--177, 2010.

\bibitem{Seth2013ProbabilisticAA}
Sohan Seth and Manuel J.~A. Eugster.
\newblock Probabilistic archetypal analysis.
\newblock {\em Machine Learning}, 102:85--113, 2013.

\bibitem{Greenacre2022PrincipalCA}
Michael Greenacre, Patrick J.~F. Groenen, Trevor Hastie, Alfonso~Iodice
  D’Enza, Angelos~I. Markos, and Elena Tuzhilina.
\newblock Principal component analysis.
\newblock {\em Nature Reviews Methods Primers}, 2:100, 2022.

\bibitem{brunet2004metagenes}
Jean-Philippe Brunet, Pablo Tamayo, Todd~R Golub, and Jill~P Mesirov.
\newblock Metagenes and molecular pattern discovery using matrix factorization.
\newblock {\em Proceedings of the National Academy of Sciences},
  101(12):4164--4169, 2004.

\bibitem{pauca2004text}
V~Paul Pauca, J~Piper, and Robert~J Plemmons.
\newblock Text mining using non-negative matrix factorizations.
\newblock In {\em Proceedings of the 2004 SIAM International Conference on Data
  Mining}, pages 452--456. SIAM, 2004.

\bibitem{Bro1997AFN}
Rasmus Bro and Sijmen de~Jong.
\newblock A fast non‐negativity‐constrained least squares algorithm.
\newblock {\em Journal of Chemometrics}, 11:393--401, 1997.

\bibitem{morch1995archetypal}
Niels M{\o}rch, Lars~Kai Hansen, Claus Svarer, and Benny Lautrup.
\newblock Archetypal analysis for machine learning and data mining.
\newblock {\em Neurocomputing}, 9(1):33--46, 1995.

\bibitem{Massey1951TheKT}
Frank~J. Massey.
\newblock The kolmogorov-smirnov test for goodness of fit.
\newblock {\em Journal of the American Statistical Association}, 46:68--78,
  1951.

\bibitem{Ross2017}
Amanda Ross and Victor~L. Willson.
\newblock {\em Independent Samples T-Test}, pages 13--16.
\newblock SensePublishers, Rotterdam, 2017.

\bibitem{Kinney2013EquitabilityMI}
Justin~B. Kinney and Gurinder~S. Atwal.
\newblock Equitability, mutual information, and the maximal information
  coefficient.
\newblock {\em Proceedings of the National Academy of Sciences}, 111:3354 --
  3359, 2013.

\bibitem{Kullback1951OnIA}
Solomon Kullback and R.~A. Leibler.
\newblock On information and sufficiency.
\newblock {\em Annals of Mathematical Statistics}, 22:79--86, 1951.

\bibitem{MENENDEZ1997307}
M.L. Men\'endez, J.A. Pardo, L.~Pardo, and M.C. Pardo.
\newblock The jensen-shannon divergence.
\newblock {\em Journal of the Franklin Institute}, 334(2):307--318, 1997.

\bibitem{Villani2009}
C{\'e}dric Villani.
\newblock {\em The Wasserstein distances}, pages 93--111.
\newblock Springer Berlin Heidelberg, Berlin, Heidelberg, 2009.

\bibitem{mmd}
Ilya Tolstikhin, Bharath~K. Sriperumbudur, and Bernhard Sch\"{o}lkopf.
\newblock Minimax estimation of maximum mean discrepancy with radial kernels.
\newblock In {\em Proceedings of the 30th International Conference on Neural
  Information Processing Systems}, NIPS'16, page 1938–1946, Red Hook, NY,
  USA, 2016. Curran Associates Inc.

\bibitem{pedregosa2011scikit}
Fabian Pedregosa, Ga{\"e}l Varoquaux, Alexandre Gramfort, Vincent Michel,
  Bertrand Thirion, Olivier Grisel, Mathieu Blondel, Gilles Louppe, Peter
  Prettenhofer, Ron Weiss, Ron~J. Weiss, J.~Vanderplas, Alexandre Passos, David
  Cournapeau, Matthieu Brucher, Matthieu Perrot, and E.~Duchesnay.
\newblock Scikit-learn: Machine learning in python.
\newblock {\em J. Mach. Learn. Res.}, 12:2825--2830, 2011.

\bibitem{McInnes2018}
Leland McInnes, John Healy, Nathaniel Saul, and Lukas Großberger.
\newblock Umap: Uniform manifold approximation and projection.
\newblock {\em Journal of Open Source Software}, 3(29):861, 2018.

\bibitem{OpenML2013}
Joaquin Vanschoren, Jan~N. van Rijn, Bernd Bischl, and Luis Torgo.
\newblock Openml: Networked science in machine learning.
\newblock {\em SIGKDD Explorations}, 15(2):49--60, 2013.

\bibitem{statlog}
Hans Hofmann.
\newblock {Statlog (German Credit Data)}.
\newblock UCI Machine Learning Repository, 1994.
\newblock {DOI}: https://doi.org/10.24432/C5NC77.

\bibitem{Harrison1978HedonicHP}
David Harrison and Daniel~L. Rubinfeld.
\newblock Hedonic housing prices and the demand for clean air.
\newblock {\em Journal of Environmental Economics and Management}, 5:81--102,
  1978.

\bibitem{ogb}
Weihua Hu, Matthias Fey, Marinka Zitnik, Yuxiao Dong, Hongyu Ren, Bowen Liu,
  Michele Catasta, and Jure Leskovec.
\newblock Open graph benchmark: datasets for machine learning on graphs.
\newblock In {\em Proceedings of the 34th International Conference on Neural
  Information Processing Systems}, NIPS '20, Red Hook, NY, USA, 2020. Curran
  Associates Inc.

\bibitem{Wu2017MoleculeNetAB}
Zhenqin Wu, Bharath Ramsundar, Evan~N. Feinberg, Joseph Gomes, Caleb Geniesse,
  Aneesh~S. Pappu, Karl Leswing, and Vijay~S. Pande.
\newblock Moleculenet: a benchmark for molecular machine learning.
\newblock {\em Chemical Science}, 9:513 -- 530, 2017.

\bibitem{Mendez2018ChEMBLTD}
David Mendez, Anna Gaulton, A.~Patr{\'i}cia Bento, Jon Chambers, Marleen
  de~Veij, Eloy Felix, Mar{\'i}a~P. Magari{\~n}os, Juan~F Mosquera, Prudence
  Mutowo, Michal Nowotka, Mar{\'i}a Gordillo-Mara{\~n}{\'o}n, Fiona M.~I.
  Hunter, Laura Junco, Grace Mugumbate, Milagros Rodr{\'i}guez-L{\'o}pez,
  Francis Atkinson, Nicolas Bosc, Chris~J. Radoux, Aldo Segura-Cabrera, Anne
  Hersey, and Andrew~R Leach.
\newblock Chembl: towards direct deposition of bioassay data.
\newblock {\em Nucleic Acids Research}, 47:D930 -- D940, 2018.

\bibitem{Yang2020MedMNISTCD}
Jiancheng Yang, Rui Shi, and Bingbing Ni.
\newblock Medmnist classification decathlon: A lightweight automl benchmark for
  medical image analysis.
\newblock {\em 2021 IEEE 18th International Symposium on Biomedical Imaging
  (ISBI)}, pages 191--195, 2020.

\bibitem{Kermany2018IdentifyingMD}
Daniel~S. Kermany, Daniel~S. Kermany, Michael~H. Goldbaum, Wenjia Cai, Carolina
  C.~S. Valentim, Huiying Liang, Sally~L. Baxter, Alex McKeown, Ge~Yang,
  Xiaokang Wu, Fangbing Yan, Justin Dong, Made~K. Prasadha, Jacqueline Pei,
  Jacqueline Pei, Magdalene Yin~Lin Ting, Jie Zhu, Christina~M. Li, Sierra
  Hewett, Sierra Hewett, Jason Dong, Ian Ziyar, Alexander Shi, Runze Zhang,
  Lianghong Zheng, Rui Hou, William Shi, Xin Fu, Xin Fu, Yaou Duan, Viet
  Anh~Nguyen Huu, Viet Anh~Nguyen Huu, Cindy Wen, Edward Zhang, Edward Zhang,
  Charlotte~L. Zhang, Charlotte~L. Zhang, Oulan Li, Oulan Li, Xiaobo Wang,
  Michael~A. Singer, Xiaodong Sun, Jie Xu, Ali~R. Tafreshi, M.~Anthony Lewis,
  Huimin Xia, and Kang Zhang.
\newblock Identifying medical diagnoses and treatable diseases by image-based
  deep learning.
\newblock {\em Cell}, 172:1122--1131.e9, 2018.

\bibitem{Yang2021MedMNISTV}
Jiancheng Yang, Rui Shi, D.~Wei, Zequan Liu, Lin Zhao, Bilian Ke, Hanspeter
  Pfister, and Bingbing Ni.
\newblock Medmnist v2 - a large-scale lightweight benchmark for 2d and 3d
  biomedical image classification.
\newblock {\em Scientific Data}, 10:41, 2021.

\bibitem{Subramanian2016ComputationalMO}
Govindan Subramanian, Bharath Ramsundar, Vijay~S. Pande, and Rajiah~Aldrin
  Denny.
\newblock Computational modeling of $\beta$-secretase 1 (bace-1) inhibitors
  using ligand based approaches.
\newblock {\em Journal of chemical information and modeling}, 56 10:1936--1949,
  2016.

\bibitem{Quirs2018UsingSS}
Miguel Quir{\'o}s, Saulius Gra{\v z}ulis, Saul\.e Girdzijauskait\.e, Andrius
  Merkys, and Antanas Vaitkus.
\newblock Using smiles strings for the description of chemical connectivity in
  the crystallography open database.
\newblock {\em Journal of Cheminformatics}, 10:23, 2018.

\bibitem{deepchem}
Bharath Ramsundar, Peter Eastman, Patrick Walters, Vijay Pande, Karl Leswing,
  and Zhenqin Wu.
\newblock {\em Deep Learning for the Life Sciences}.
\newblock O'Reilly Media, 2019.
\newblock
  \url{https://www.amazon.com/Deep-Learning-Life-Sciences-Microscopy/dp/1492039837}.

\bibitem{Rupp2011FastAA}
Matthias Rupp, Alexandre Tkatchenko, Klaus-Robert M{\"u}ller, and O.~Anatole
  von Lilienfeld.
\newblock Fast and accurate modeling of molecular atomization energies with
  machine learning.
\newblock {\em Physical review letters}, 108 5:058301, 2011.

\bibitem{metabolomics}
Manish Sud, Eoin Fahy, Dawn Cotter, Kenan Azam, Ilango Vadivelu, Charles
  Burant, Arthur Edison, Oliver Fiehn, Richard Higashi, K.~Sreekumaran Nair,
  Susan Sumner, and Shankar Subramaniam.
\newblock Metabolomics workbench: An international repository for metabolomics
  data and metadata, metabolite standards, protocols, tutorials and training,
  and analysis tools.
\newblock {\em Nucleic Acids Research}, 44(D1):D463--D470, 10 2015.

\bibitem{ghg}
D.~Lucas.
\newblock {Greenhouse Gas Observing Network}.
\newblock UCI Machine Learning Repository, 2015.
\newblock {DOI}: https://doi.org/10.24432/C5JK5M.

\bibitem{Dong2024OnlineMG}
Haiqi Dong, Amanda~S. Barnard, and Amanda~J Parker.
\newblock Online meta-learned gradient norms for active learning in science and
  technology.
\newblock {\em Machine Learning: Science and Technology}, 5:015041, 2024.

\bibitem{coresets}
Yu~Yang, Hao Kang, and Baharan Mirzasoleiman.
\newblock Towards sustainable learning: coresets for data-efficient deep
  learning.
\newblock In {\em Proceedings of the 40th International Conference on Machine
  Learning}, ICML'23. JMLR.org, 2023.

\end{thebibliography}

\end{document}